\documentclass[letterpaper]{article} 
\usepackage{aaai24}  
\usepackage{times}  
\usepackage{helvet}  
\usepackage{courier}  
\usepackage[hyphens]{url}  
\usepackage{graphicx} 
\usepackage{natbib}  
\usepackage{caption} 
\usepackage{algorithm}
\usepackage{algorithmic}

\usepackage{amssymb}
\usepackage{amsmath}
\usepackage{multirow}
\usepackage{caption}
\usepackage{subcaption}
\usepackage{pifont}
\usepackage{makecell}

\usepackage{newfloat}
\usepackage{listings}
\DeclareCaptionStyle{ruled}{labelfont=normalfont,labelsep=colon,strut=off} 
\floatstyle{ruled}
\newfloat{listing}{tb}{lst}{}
\floatname{listing}{Listing}
\title{Out-of-Distribution Detection in Long-Tailed Recognition \\with Calibrated Outlier Class Learning}
\author{
    Wenjun Miao\textsuperscript{\rm 1},
    Guansong Pang\textsuperscript{\rm 2*},
    Xiao Bai\textsuperscript{\rm 1,\rm 3},
    Tianqi Li\textsuperscript{\rm 1},
    Jin Zheng\textsuperscript{\rm 1, \rm 4}\thanks{Corresponding authors: 
    G. Pang (gspang@smu.edu.sg) and 
    J. Zheng (jinzheng@buaa.edu.cn)}
}
\affiliations{
    \textsuperscript{\rm 1}School of Computer Science and Engineering, Beihang University\\
    \textsuperscript{\rm 2}School of Computing and Information Systems, Singapore Management University\\
    \textsuperscript{\rm 3}State Key Laboratory of Software Development Environment, Jiangxi Research Institute, Beihang University\\ 
    \textsuperscript{\rm 4}State Key Laboratory of Virtual Reality Technology and Systems, Beihang University\\
}

\usepackage{bibentry}

\begin{document}

\maketitle

\begin{abstract}
Existing out-of-distribution (OOD) methods have shown great success on balanced datasets but become ineffective in long-tailed recognition (LTR) scenarios where 1) OOD samples are often wrongly classified into head classes and/or 2) tail-class samples are treated as OOD samples. To address these issues, current studies fit a prior distribution of auxiliary/pseudo OOD data to the long-tailed in-distribution (ID) data. However, it is difficult to obtain such an accurate prior distribution given the unknowingness of real OOD samples and heavy class imbalance in LTR. A straightforward solution to avoid the requirement of this prior is to learn an outlier class to encapsulate the OOD samples. The main challenge is then to tackle the aforementioned confusion between OOD samples and head/tail-class samples when learning the outlier class. To this end, we introduce a novel calibrated outlier class learning (COCL) approach, in which 1) a debiased large margin learning method is introduced in the outlier class learning to distinguish OOD samples from both head and tail classes in the representation space and 2) an outlier-class-aware logit calibration method is defined to enhance the long-tailed classification confidence. Extensive empirical results on three popular benchmarks CIFAR10-LT, CIFAR100-LT, and ImageNet-LT demonstrate that COCL substantially outperforms state-of-the-art OOD detection methods in LTR while being able to improve the classification accuracy on ID data. Code is available at \url{https://github.com/mala-lab/COCL}.
\end{abstract}

\begin{figure}[t!]
  \centering
  \subfloat[Feature representations of CIFAR100-LT test data]
  {\includegraphics[width=1\columnwidth]{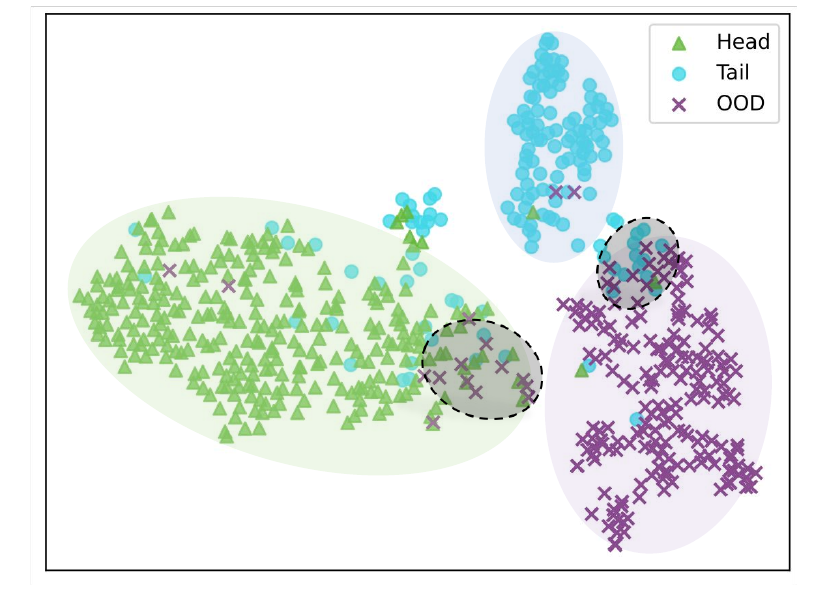} \label{t-SNE}} \\
  \subfloat[Prediction confidence]
  {\includegraphics[width=0.50\columnwidth]{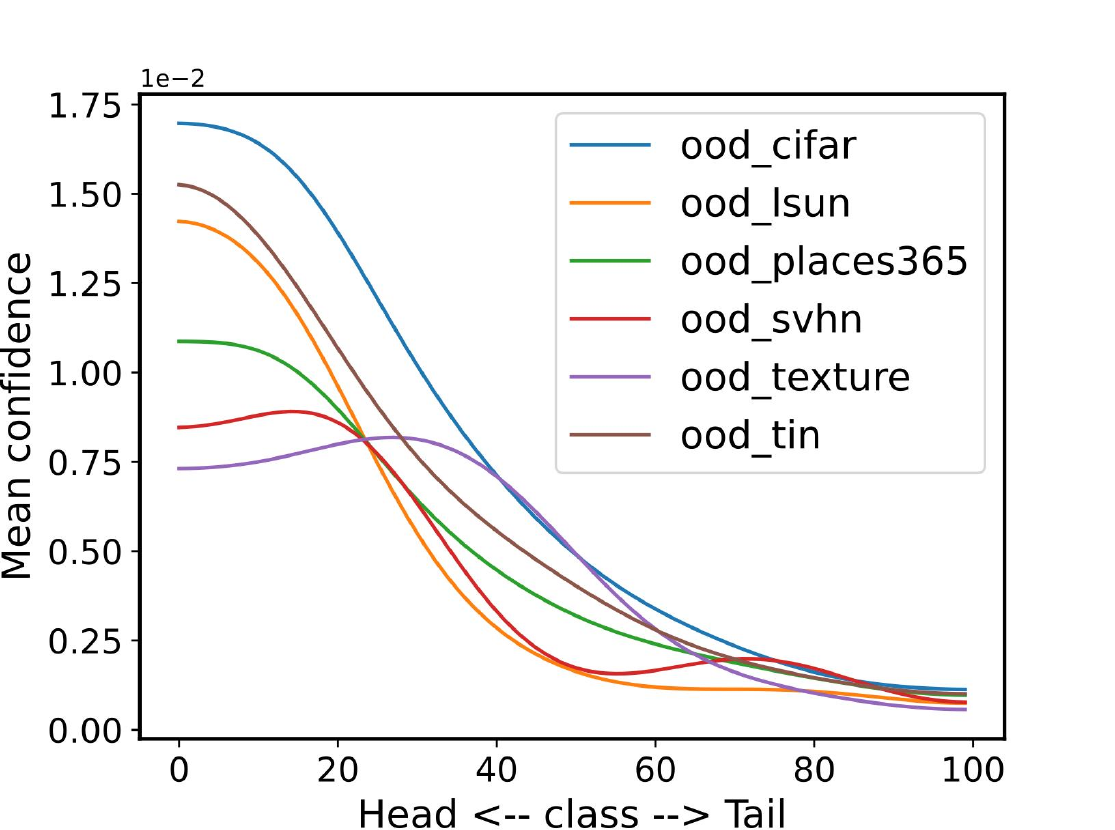}\label{OOD2ID_guass}}
  \subfloat[OOD score for ID samples]
  {\includegraphics[width=0.50\columnwidth]{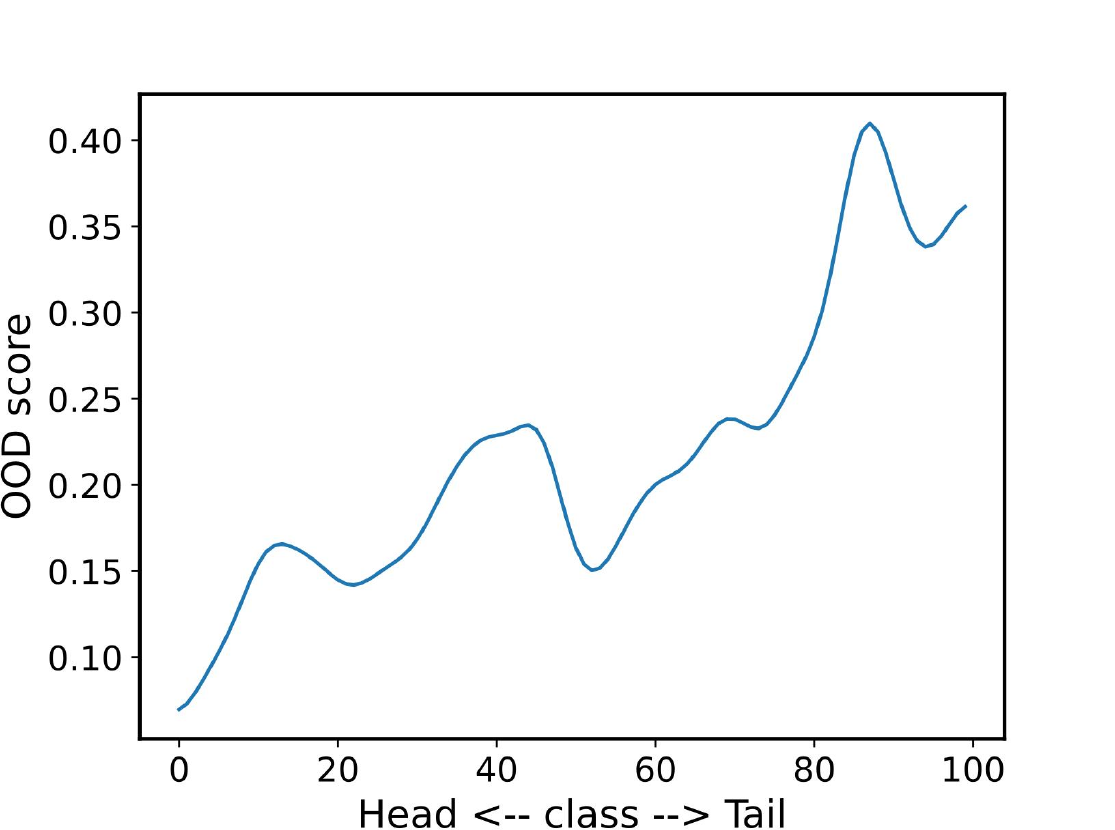}\label{ID2OOD}}
  \caption{Visualization and qualitative results on test data of CIFAR100-LT using an LTR model augmented with an outlier learning module (see Eq. \ref{OCL}) for OOD detection. (a) Feature representations of samples randomly selected from the head class, tail class, and OOD samples. The gray areas highlight obscure regions between head/tail samples and OOD samples. (b) The mean prediction confidence of the model classifying six OOD datasets into one of the ID classes. (c) The mean OOD score (i.e., the softmax probability of the outlier class) of samples from each ID class.}
  \label{motivate}
\end{figure}

\section{Introduction}

Deep neural networks (DNNs) have achieved remarkable success across various fields \cite{russakovsky2015imagenet, krizhevsky2017imagenet}. However, their application in real-world scenarios, such as autonomous driving \cite{kendall2017uncertainties} and medical diagnosis \cite{leibig2017leveraging}, remains challenging due to the presence of long-tailed distribution and unknown classes \cite{huang2021mos, wang2020long}. 
In particular, DNNs often have high confidence predictions that classify out-of-distribution (OOD) samples from unknown classes as one of the known classes. This issue is further amplified when the in-distribution (ID) data has a class-imbalanced/long-tailed distribution \cite{zhu2023openmix, li2022targeted, li2022long,wang2022partial}. 
This is because, as illustrated in Fig. \ref{t-SNE}, DNNs trained on long-tailed data can be heavily biased towards head classes (the majority classes) due to the overwhelming presence of samples from these classes, and as a result, the long-tailed recognition (LTR) models often misclassify OOD samples into the head classes with high confidence (see Fig. \ref{OOD2ID_guass}); further, the LTR models 
tend to treat tail samples as part of OOD samples due to the rareness of tail samples in the training data, i.e., the tail samples often have a much higher OOD score than the head samples (see Fig. \ref{ID2OOD}). 

Compared to OOD detection on balanced ID datasets, significantly less work has been done in the LTR scenarios. Recent studies \cite{wang2022partial, wei2022open, jiang2023detecting, choi2023balanced} are among the seminal works exploring OOD detection in LTR. The current methods in this line focus on distinguishing OOD samples from ID samples by an approach called outlier exposure (OE) \cite{hendrycks2018deep} that fits auxiliary/pseudo OOD data to a prior distribution (e.g., uniform distribution) of ID data. However, unlike balanced ID datasets, LTR datasets heavily skewed the distribution of ID data, so using the commonly-adopted uniform distribution as the prior becomes ineffective. Estimating this prior from the sample size of ID classes is a simple solution to alleviate this issue, but it can intensify the LTR models' bias toward head classes.
Another line of approach is focused on learning discriminative representations to separate OOD samples from tail samples. However, the lack of sufficient samples in the tail classes renders this approach less effective, furthermore, it often fails to distinguish head and OOD samples.

In this work, we aim to synthesize both approaches and introduce a novel approach, namely calibrated outlier class learning (COCL). Intuitively, a straightforward solution to avoid the requirement of this prior in the OE-based approach is to learn an outlier class to encapsulate the OOD samples. The main challenge is then mainly about tackling the aforementioned confusion between OOD samples and head/tail-class samples when learning the outlier class. To address this challenge, we introduce a debiased large margin learning method, which is jointly optimized with the outlier class learning to distinguish OOD samples from both head and tail classes in the representation space. We further introduce an outlier-class-aware logit calibration method that takes into account the outlier class when calibrating the ID prediction probability. This helps enhance the long-tailed classification confidence while improving the OOD detection performance.
In summary, our main contributions are as follows:
\begin{itemize}
\item We show that outlier class learning is generally more effective for OOD detection in LTR than fitting to a prior distribution when auxiliary OOD data is available.
\item We then introduce a novel calibrated outlier class learning (COCL) approach that learns an accurate LTR model with a strong OOD detector that effectively mitigates the biases towards head and OOD samples. To this end, we introduce two components, including the debiased large margin learning and the outlier-class-aware logit calibration, which work in the respective training and inference stages, 
enabling substantially improved OOD detection and long-tailed classification performance. 
\item 
Extensive empirical results on three popular benchmarks CIFAR10-LT, CIFAR100-LT, and ImageNet-LT demonstrate that COCL substantially outperforms state-of-the-art OOD detection methods in LTR while improving the classification accuracy on ID data.
\end{itemize}

\section{Related Work}
\subsubsection{OOD Detection}
The objective of this task is to determine whether a given input sample belongs to known classes (in-distribution) or unknown classes (out-of-distribution). 
In recent years, OOD detection has been developed extensively, including post-hoc strategies \cite{sun2021react, wang2023detecting, zhang2023decoupling} and training-time strategies \cite{liu2020energy, wei2022mitigating, tian2022pixel,yu2023block, li2023rethinking,liu2023residual}. 
The post-hoc methods focus on devising new OOD scoring functions in the inference phase.
, such as MSP \cite{hendrycks2016baseline}, Mahalanobis distance \cite{lee2018simple}, Gram matrix \cite{sastry2020detecting}. 
The training-time methods focus on separating OOD samples from ID samples by utilizing auxiliary data during training. 
Outlier exposure (OE) \cite{hendrycks2018deep} is arguably the most popular approach in this line that utilizes the OOD data by enforcing a uniform distribution of its prediction probability to ID classes. 
EnergyOE \cite{liu2020energy} improves OE and maximizes the free energy of OOD samples instead. 
UDG \cite{yang2021semantically} introduces unsupervised dual grouping to leverage unlabelled auxiliary data for OOD detection. 
However, all these methods are focused on cases with balanced ID training data, which fail to work well on imbalanced ID datasets.

\subsubsection{Long-Tailed Recognition (LTR)}
LTR aims at improving the accuracy of tail classes with the least influence on the head classes. Re-sampling 
\cite{wang2020devil, tang2022invariant, bai2023effectiveness}
and re-weighting \cite{tan2020equalization, alshammari2022long, gou2023rethinking, hong2023long} 
that focus on balancing the ratio between head and tail classes are the most straightforward solutions for LTR. 
Two-stage methods \cite{kang2019decoupling, nam2023decoupled} is another approach that retrains the classifier on a re-balanced dataset during the fine-tuning stage, leading to significant improvement in LTR. 
Additionally, logit adjustment (LA) \cite{menon2020long} emerges as an effective statistical framework that can be applied in both the training and inference phases to further enhance the ID recognition performance. 
Although these LTR methods show effective performance in the long-tailed classification of ID samples, they do not have an explicit design to handle OOD samples.

\begin{table*}[t!]
\caption{Comparison of outlier exposure (OE) and outlier class learning (OCL) approaches when combined with three LTR methods. 
All methods are trained on CIFAR10/100-LT using 
ResNet18. Reported are the average performance across six
different OOD test sets (including CIFAR, Texture, SVHN, LSUN, Places365, and TinyImagenet) in the commonly-used SC-OOD detection benchmark \cite{yang2021semantically}
(See the Experiments section for the description of evaluation measures).
}
\centering
\footnotesize
\scalebox{0.85}{
\begin{tabular}{ccc|ccccc|ccccc}
\hline
\multirow{2}{1cm}{\centering OOD \\ Method} &  & \multirow{2}{1cm}{\centering LTR \\ Method} & \multicolumn{5}{c|}{CIFAR10-LT} & \multicolumn{5}{c}{CIFAR100-LT} \\ \cline{4-13}
 & &  & AUC$\uparrow$ & AP-in$\uparrow$ & AP-out$\uparrow$ & FPR$\downarrow$ & ACC$\uparrow$ & AUC$\uparrow$ & AP-in$\uparrow$ & AP-out$\uparrow$ & FPR$\downarrow$ & ACC$\uparrow$ \\
\hline
\multirow{4}{*}{OE} & \multirow{4}{*}{\makebox[0.01\textwidth][c]{+}} & None (baseline) & 89.76 & 89.45 & 87.22 & 53.19 & 73.59 & 73.52 & 75.06 & 67.27 & 86.30 & 39.42 \\
& & Re-weight & 89.34 & 88.63 & 86.39 & 56.24 & 70.35 & 73.08 & 73.86 & 66.05 & 87.22 & 39.45 \\
& & $\tau$-norm & 89.58 & 88.21 & 85.88 & 52.84 & 73.33 & 73.62 & 74.67 & 66.59 & 86.02 & 40.87 \\
& & LA & 89.46 & 88.74 & 86.39 & 53.38 & 73.93 & 73.44 & 74.33 & 66.48 & 86.13 & 42.06 \\ 
\hline
\multirow{4}{*}{\shortstack{Outlier class \\ learning}} & \multirow{4}{*}{\makebox[0.01\textwidth][c]{+}} & None (baseline) & 89.91 & 88.15 & 90.38 & 41.13 & 74.48 & 73.56 & 74.12 & 69.65 & 81.93 & 41.54 \\
& & Re-weight & 90.45 & 89.12 & 90.58 & 38.86 & 74.84 & 74.23 & 74.29 & 70.68 & 79.45 & 42.06 \\
& & $\tau$-norm & 90.95 & 89.59 & 91.11 & 37.91 & 75.14 & 74.57 & 75.12 & 70.76 & 81.27 & 44.21 \\
& & LA & 91.56 & 90.52 & 91.51 & 36.50 & 76.67 & 74.77 & 75.15 & 71.13 & 80.33 & 43.02 \\ 
\hline
\multicolumn{3}{c|}{Our method COCL} & \textbf{93.28} & \textbf{92.24} & \textbf{92.89} & \textbf{30.88} & \textbf{81.56} & \textbf{78.25} & \textbf{79.37} & \textbf{73.58} & \textbf{74.09} & \textbf{46.41} \\
\hline
\end{tabular}
}
\label{outlier_framework}
\end{table*}

\subsubsection{OOD Detection in LTR}
PASCL \cite{wang2022partial} formulates the OOD detection problem in LTR and reveals the difficulty that simple combinations of existing OOD detection and LTR methods do not work well.
Particularly, PASCL evaluates different baseline methods in the SC-OOD benchmark \cite{yang2021semantically} to establish performance benchmarks for OOD detection in LTR. 
Open sampling \cite{wei2022open} finds that leveraging equivalent noisy labels does not harm the training, so it introduces a noisy labels assignment method for utilizing unlabeled auxiliary OOD data to enhance the robustness of OOD detection and improve ID classification accuracy. 
Recent studies \cite{choi2023balanced, jiang2023detecting} find that fitting the prediction probability of OOD data to a long-tailed distribution in either the scratch or fine-tuning approach is more effective than using a uniform distribution. 
They specify this prior distribution based on the number of samples in ID classes or a pre-trained ID model to learn the OOD detection model. 
However, it is difficult to obtain such an accurate prior distribution of OOD data in LTR. We instead utilize the outlier class learning to eliminate the need for such a prior.

\section{Outlier Class Learning vs. Outlier Exposure}
\subsubsection{Problem Statement}
Let $\mathcal{X} = X^{in} \cup X^{out}$ denote the input space and $Y^{in} = \left\{ 1,2,\dots,k \right\}$ be the set of $k$ imbalanced ID classes in the label space. OOD detection in LTR is to learn a classifier $f$ that for any test data $x \in \mathcal{X}$: if $x$ drawn from $X^{in}$ (from either head or tail classes), then $f$ can classify $x$ into correct ID class; and if $x$ is drawn from $X^{out}$, then $f$ can detect $x$ as OOD data. 
It is normally assumed that genuine OOD data $X^{out}$ is not available during training since OOD samples are unknown instances. On the other hand, auxiliary samples that are not $X^{out}$ but drawn from a different distribution other than $X^{in}$ are often available. These auxiliary samples can be used as pseudo OOD samples to fine-tune/re-train the LTR models.

\subsubsection{Outlier Exposure (OE)}
OE is a popular OOD detection approach that uses the auxiliary data as outliers to train ID classifiers for separating ID and OOD samples.
Specifically, given ID data $\mathcal{D}_{in} = (X_{in}, Y_{in})$ and auxiliary data $\mathcal{D}_{out} = (X_{out}, u)$ for training, where $u$ is a uniform distribution-based pseudo label for OOD data, OE then minimizes:
\begin{equation}
\mathcal {L}_{OE} =  \mathbb {E}_{x,y \sim \mathcal {D}_{in}}[\ell(f(x),y] + \gamma \mathbb {E}_{x \sim \mathcal {D}_{out}}[\ell(f(x),u], \label{OE}
\end{equation}
where $ \gamma $ denotes a hyper-parameter, and $ \ell $ is a cross entropy loss. During inference, it uses the maximum softmax probability (MSP) over the ID classes as an OOD score.

\subsubsection{Outlier Class Learning (OCL)}
Outlier class learning (OCL) aims at learning a new (outlier) class that encapsulates the OOD samples, rather than enforcing a uniform prediction probability distribution as in the second term of Eq. \ref{OE}. Specifically, for a $ k $-class classification problem, it extends the label space by explicitly adding a separate class $k+1$ as outlier class, i.e., ID data $\mathcal{D}_{in} = (X_{in}, Y_{in})$ and auxiliary data $\mathcal{D}_{out} = (X_{out}, k+1)$ are used during training, and we then minimize the following loss function:
\begin{equation}
\mathcal {L}_{OCL} =  \mathbb {E}_{x,y \sim \mathcal {D}_{in}}[\ell(f(x),y] + \gamma \mathbb {E}_{x \sim \mathcal {D}_{out}}[\ell(f(x),\tilde{y}], \label{OCL}
\end{equation}
where $ \tilde{y}=k+1 $ and $ \gamma $ denotes a hyper-parameter. The softmax probability from the $k+1$ class is used as an OOD score during inference.

\subsubsection{OCL Aligns Better with LTR Than OE}
We find empirically that OE achieves promising performance in general OOD detection scenarios, but works less effectively when applied to LTR settings. It is mainly because the uniform prediction probability prior in Eq. \ref{OE} does not hold in LTR. OCL helps eliminate this prior input and learns the outlier class that separates the OOD samples from the ID samples in the representation space.
In Table \ref{outlier_framework}, we compare the performance of OE and OCL when combining with three widely used LTR methods: Re-weight \cite{cui2019class}, $\tau$-norm \cite{kang2019decoupling}, and logit adjustment (LA) \cite{menon2020long}. 
The results show that OCL largely improves not only ID classification accuracy but also OOD detection performance on both datasets, substantially outperforming the OE method which is explored for LTR settings in PASCL \cite{wang2022partial}. Motivated by the large performance gap, we promote the use of OCL for LTR instead. However, there are two main challenges in the OCL approach: 1) OOD samples can often be wrongly classified into head classes and/or 2) tail-class samples are often misclassified as OOD samples. Our approach calibrated OCL (COCL) is focused on addressing these two challenges, and as shown in Table \ref{outlier_framework}, 
it can help address the challenges and achieve largely improved classification and detection performance over the general OCL baselines.

\begin{figure*}[t!]
  \centering
  \subfloat[Pipeline]
  {\includegraphics[width=0.48\textwidth]{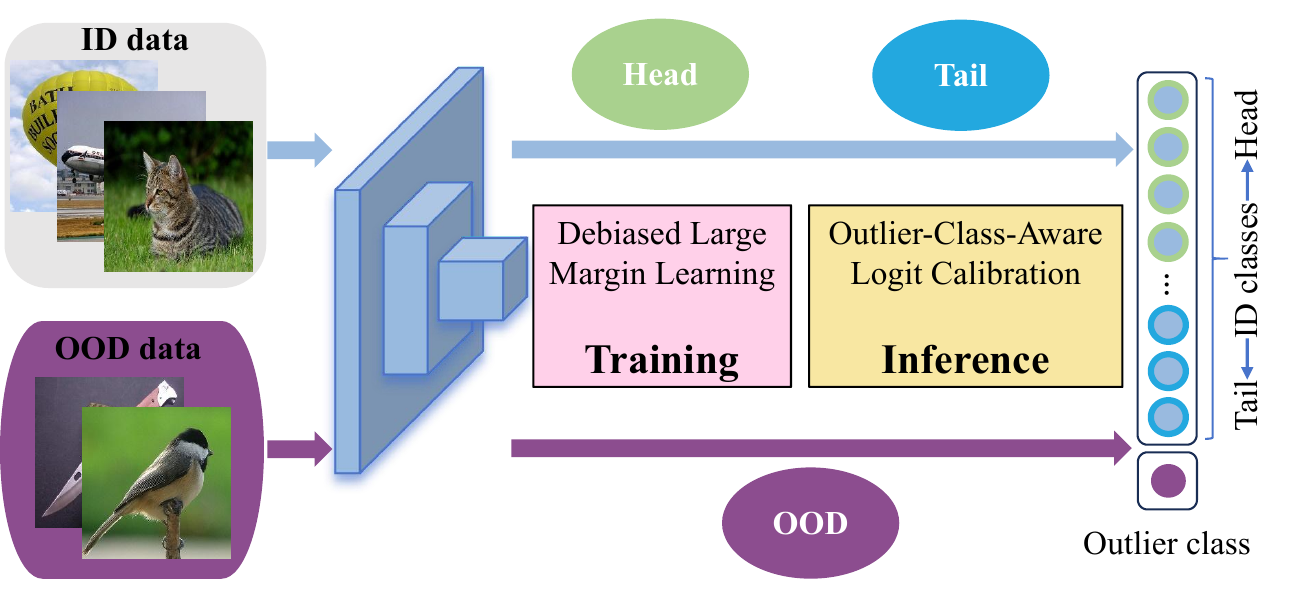} \label{pipeline}} 
  \subfloat[Debiased large margin learning]
  {\includegraphics[width=0.35\textwidth]{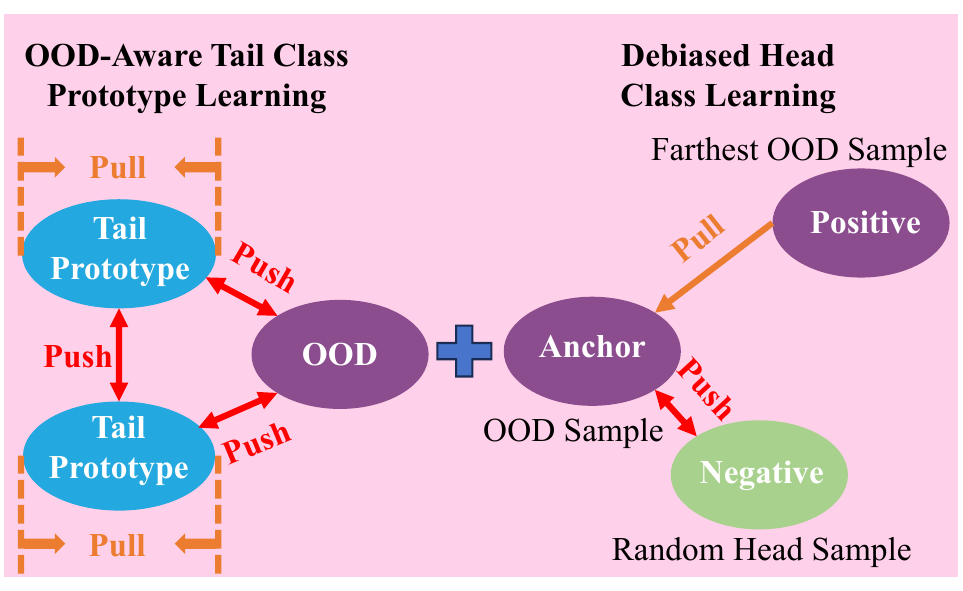}\label{debias_large_margin}} 
  \subfloat[Outlier-class-aware logit calibration]
  {\includegraphics[width=0.17\textwidth]{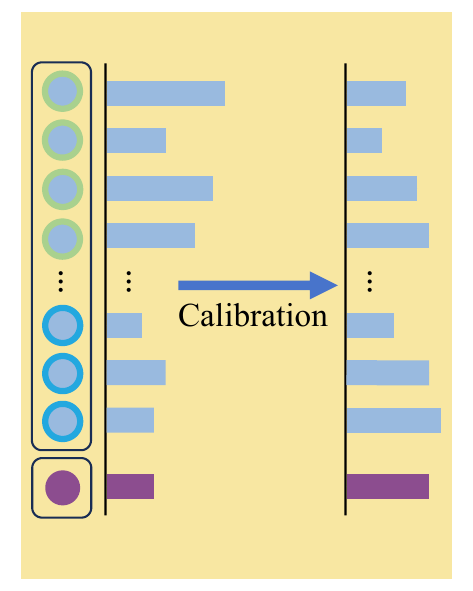}\label{logit_calibration}}
  \caption{Overview of our approach COCL. (a) presents a high-level pipeline of our two components in our approach COCL, (b) illustrates the key idea of debiased large margin learning which includes OOD-aware tail class prototype learning and Debiased Head Class Learning to reduce biases towards OOD samples and head classes respectively, and (c) shows the outlier-class-aware logit calibration that utilizes the logit of the outlier class to calibrate the prediction results during inference.}
  \label{overview}
\end{figure*}

\section{Approach}

\subsection{Overview of Our Proposed Approach COCL}
We introduce a novel COCL approach to tackle the aforementioned two issues for OOD detection in LTR. COCL consists of two components, namely \textit{debiased large margin learning} and 
\textit{outlier-class-aware logit calibration}, as shown in Fig. \ref{overview}. The debiased large margin learning, as shown in Fig. \ref{debias_large_margin}, is designed to reduce the bias towards head classes (leading to the misclassification of OOD samples into head classes) as well as the bias towards OOD samples (leading to the misclassification of tail samples as OOD samples) during training.
The outlier-class-aware logit calibration component, as shown in Fig. \ref{logit_calibration}, is devised to utilize the logit of the outlier class for calibration to enhance OOD detection and the confidence of long-tailed classification during inference.
Below we introduce each component in detail.

\subsection{Debiased Large Margin Learning}
The debiased large margin learning component includes two modules, namely OOD-aware tail class prototype learning and Debiased Head Class Learning, to respectively reduce the model bias towards OOD samples and head classes. Below we elaborate on how these two modules can help reduce the two types of model bias.

\subsubsection{OOD-Aware Tail Class Prototype Learning}
Since tail class samples are rare in the training data, the LTR models lack confidence in classifying them. As a result, they tend to exhibit high OOD scores during LTR inference, i.e., the LTR models' bias towards OOD samples when classifying tail class samples. 
The seminal work PASCL \cite{wang2022partial} 
attempts to utilize diverse augmentations to push tail samples away from OOD samples, but it often learns non-discriminative representations between OOD samples and tail samples due to the limited size of tail classes. 
To address this issue, we utilize a learnable prototype of one tail class as positive sample to pull tail samples closer to their prototype, 
with OOD samples and other tail class prototypes as negative samples to push the samples and prototype of the positive tail class away from OOD samples and other tail prototypes.
This strategy harnesses the tail prototypes to increase the presence of representations for tail classes, helping reduce the model bias towards the OOD samples.
Formally, let $\mathcal {M} \in \mathbb R^{N \times D}$ be the learnable parameters of $N$ tail prototypes, with each prototype representation spanned in a $D$-dimensional space, our tail class prototypes are learned by minimizing the following loss:\\
\begin{equation}
    \mathcal {L}_{t}=\mathbb E_{x \sim \mathcal {D}_{tail}}[\mathcal {L}_{t}(x, \mathcal {M})] ,
\end{equation} 
where $\mathcal{D}_{tail}$ is all tail samples in $\mathcal{D}_{in}$,  and $ \mathcal {L}_{t}(x, \mathcal {M})$ is defined as:
\begin{equation}
\mathcal {L}_{t}(x, \mathcal {M})= \frac{1}{\left| \mathcal B \right|} \sum_{x \in \mathcal B}  log \frac{exp(z(x)m_x^{\intercal}/t)}{\sum\limits_{m \in \mathcal M}exp(z(x)m^{\intercal} / t) + P(x)}, \label{OODT}
\end{equation}
where $\mathcal{B}$ is a training sample batch, $z(\cdot)$ is the output of a non-linear projection on the model’s penultimate layer, i.e., the learned feature representation of $x$, $P(x) =\sum_{\hat{x} \in \mathcal O} exp(z(x)z(\hat{x})^{\intercal}/t)$ with $\mathcal{O}$ being a batch of OOD samples from $\mathcal{D}_{out}$, $\intercal$ is a transpose operation,
$ m $ is a tail prototype in $\mathcal {M}$, $ m_x $ is the tail prototype corresponding to the tail class of sample $x$, and $t$ is the scaling temperature. 

As illustrated on the left in Fig. \ref{debias_large_margin}, we only apply this tail class prototype learning to the tail-class in-distribution data and OOD data, as it is specifically designed to tackle the problem that tail samples exhibit high OOD scores. 
As for the head samples, they are normally easily distinguished from the OOD samples as there are sufficient head class samples in the training set.
Note that we exclusively calculate the loss only when taking tail samples as input; the loss is not calculated for auxiliary OOD samples, since we have already pulled OOD samples together in the joint LTR and outlier class learning in Eq. \ref{OCL}. 
Thus, this module introduces only minor computation overheads to the general OCL.

\subsubsection{Debiased Head Class Learning}
Due to the overwhelming presence of head class samples, LTR models demonstrate a strong bias towards head classes when performing OOD detection, i.e., OOD samples are often misclassified as one of the head classes.
To address this issue, we introduce the debiased head class learning module that performs a one-class learning of OOD samples, where we aim to learn a large outlier-class description region for OOD samples to alleviate the dominant influence of head samples in the feature space. 
To this end, as illustrated on the right in Fig. \ref{debias_large_margin}, we use only the OOD samples as anchors, with randomly sampled head samples as negative samples and the OOD samples that are distant from the anchors in the feature space as positive samples, and then we perform a semi-supervised one-class learning for OOD samples by minimizing the following loss:
\begin{equation}
\mathcal {L}_{h}=\mathbb E_{x \sim \mathcal {D}_{out}}[\mathcal {L}_{h}(x)],
\end{equation}
where $ \mathcal {L}_{h}(x)$ is defined as:
\begin{align}
\begin{aligned}
 \mathcal {L}_{h}(x)=\frac{1}{\left| \mathcal B \right|} \sum_{x \in \mathcal B}  max(0, \| z(x) - z(x^p) \|_2^2 \\
- \| z(x) - z(x^n) \|_2^2 + margin), 
\label{OODH} 
\end{aligned}
\end{align}
where $\mathcal B$ is a batch of OOD training samples from $\mathcal{D}_{out}$, $o^p$ is a positive sample that is set to the most distance OOD sample from the anchor sample $x$ in $\mathcal B$, $x^n$ is a randomly head sample in the same batch, and $margin$ is a user-defined hyperparameter that specifies the margin between the one-class OOD description region and the head samples.
Note that since popular contrastive learning is a two-way learning method, the model would be reinforced to bias towards the head class if the original contrastive learning is directly applied. Our design in Eq. \ref{OODH} is to explicitly correct this bias and refine the learning of the outlier class. 

Lastly, the overall objective of our debiased large margin learning is as follows:
\begin{align}
\begin{aligned}
\mathcal {L}_{total} & = \mathcal {L}_{OCL} + \alpha \mathcal {L}_{t} + \beta \mathcal {L}_{h}  \\ 
& =\mathbb {E}_{x,y \sim \mathcal {D}_{in}}[\ell(f(x),y] + \gamma \mathbb {E}_{x \sim \mathcal {D}_{out}}[\ell(f(x),\tilde{y}] \\ 
& + \alpha \mathbb E_{x \sim \mathcal {D}_{tail}}[\mathcal {L}_{t}(x, \mathcal {M})] + \beta \mathbb E_{x \sim \mathcal {D}_{out}}[\mathcal {L}_{h}(x)],
\label{Total} 
\end{aligned}
\end{align}
where $\mathcal {L}_{OCL}$ is the same as the outlier class learning in Eq. (\ref{OCL}), $\mathcal {L}_{t}$ is as defined in Eq. (\ref{OODT}), and $\mathcal {L}_{h}$ is as defined in Eq. (\ref{OODH}). $\alpha$ and $\beta$ denote two hyperparameters to control the reduction of biases towards the OOD data and head classes.

\subsection{Outlier-Class-Aware Logit Calibration}
Our LTR model is then equipped with an OOD detector by minimizing Eq. \ref{Total} on the training data. 
However, due to the inherent class imbalance in the training data, the LTR model often tends to have a higher confidence on the prediction of head samples than both the tail samples and OOD samples.
To avoid this issue, we propose the outlier-class-aware logit calibration component that calibrates the predictions using the model logits and prior probability of both ID and outlier classes in the inference stage. This is different from existing LTR calibration methods that are focused on the ID classes only. Specifically, given a test sample $x$, we calibrate its posterior probability via:
\begin{equation}
P(y=i|x) = \frac{e^{f_i(x) - \tau \cdot \log n_i}}{\sum_{j=1}^{k+1}e^{f_j(x)-\tau \cdot \log n_j}}, \label{E7} 
\end{equation}
where $f_i(x)$ denotes the predicted model logit of $x$ belonging to the class $i$, $\tau$ is a hyperparameter to balance how much we want to bring in the prior of the outlier class, and $n_i$ is a prior probability for the class $i$ and it is estimated by
\begin{equation}
n_i = \frac{N_i}{N_1+N_2+ \cdots + N_k}, \label{E6} 
\end{equation}
where $N_i$ is the training sample size for class $i$. We do not have genuine OOD samples during training, so their prior probability can not be estimated in the same way as ID classes. Motivated by the fact that detecting OOD samples should be as important as ID classification, we set $n_{k+1}=1$ for the outlier class, which equals the summation of the prior probabilities of all ID classes.
In doing so, our model's prediction is calibrated to decrease the probability of head classes and increase that of tail classes, while taking into account the influence of OOD samples on the prediction. Thus, this calibration is beneficial to both ID classification and OOD detection. This effect cannot be achieved using the general logit calibration used in LTR.

\begin{table}[t!]
\caption{Comparison results on CIFAR10-LT.}
\begin{subtable}{1\linewidth}
\centering
\footnotesize
\caption{Comparison of COCL with OE and OCL on six OOD datasets.}
\label{cifar10_table_top}
\scalebox{0.85}{
\begin{tabular}{c|c|c|c|c|c}
\hline
OOD Dataset & Method & AUC$\uparrow$ & AP-in$\uparrow$ & AP-out$\uparrow$ & FPR$\downarrow$ \\ 
\hline
\multirow{3}{*}{Texture} & OE & 92.30 & 96.01 & 82.57 & 48.65 \\ 
 & OCL & 93.71 & 95.95 & 91.07 & 27.22 \\ \cline{2-6}
 & \textbf{COCL (Ours)} & \textbf{96.81} & \textbf{98.21} & \textbf{93.86} & \textbf{14.65} \\
\hline
\multirow{3}{*}{SVHN} & OE & 94.86 & 91.59 & 97.00 & 29.11 \\
 & OCL & 95.14 & 90.88 & 97.73 & 25.47 \\ \cline{2-6}
 & \textbf{COCL (Ours)} & \textbf{96.98} & \textbf{93.25} & \textbf{98.61} & \textbf{12.59} \\
\hline
\multirow{3}{*}{CIFAR100} & OE & 83.32 & 84.06 & 80.83 & 65.82 \\
 & OCL & 82.04 & 82.52 & 81.92 & 63.35 \\ \cline{2-6}
 & \textbf{COCL (Ours)} & \textbf{86.63} & \textbf{86.66} & \textbf{86.28} & \textbf{52.21} \\
\hline
\multirow{3}{*}{\shortstack{Tiny \\ ImageNet}} & OE & 86.35 & 89.88 & 79.30 & 64.50 \\
 & OCL & 85.90 & 88.98 & 82.17 & 57.46 \\ \cline{2-6}
 & \textbf{COCL (Ours)} & \textbf{90.43} & \textbf{92.52} & \textbf{87.03} & \textbf{46.12} \\
\hline
\multirow{3}{*}{LSUN} & OE & 91.57 & 93.06 & 88.37 & 53.99 \\
 & OCL & 92.75 & 92.69 & 93.10 & 30.95 \\ \cline{2-6}
 & \textbf{COCL (Ours)} & \textbf{94.85} & \textbf{95.43} & \textbf{93.98} & \textbf{27.48} \\
\hline
\multirow{3}{*}{Place365} & OE & 90.20 & 82.09 & 95.24 & 57.06 \\
 & OCL & 89.91 & 77.91 & 96.28 & 42.33 \\ \cline{2-6}
 & \textbf{COCL (Ours)} & \textbf{93.97} & \textbf{87.36} & \textbf{97.56} & \textbf{32.25} \\
\hline
\multirow{3}{*}{Average} & OE & 89.76 & 89.45 & 87.22 & 53.19 \\
 & Outlier & 89.91 & 88.15 & 90.38 & 41.13 \\ \cline{2-6}
 & \textbf{Ours} & \textbf{93.28} & \textbf{92.24} & \textbf{92.89} & \textbf{30.88} \\
\hline
\end{tabular}
}\\

\caption{\centering{Comparison results with different competing methods. The results are averaged over the six OOD test datasets in (a).}}
\scalebox{0.85}{
\begin{tabular}{c|c|c|c|c|c}
\hline
Method & AUC$\uparrow$ & AP-in$\uparrow$ & AP-out$\uparrow$ & FPR$\downarrow$ & ACC$\uparrow$ \\ 
\hline
MSP & 74.33 & 73.96 & 72.14 & 85.33 & 72.17\\
OE & 89.76 & 89.45 & 87.22 & 53.19 & 73.59\\
EnergyOE & 91.92 & 91.03 & 91.97 & 33.80 & 74.57\\
OCL & 89.91 & 88.15 & 90.38 & 41.13 & 74.48 \\
PASCL & 90.99 & 90.56 & 89.24 & 42.90 & 77.08\\
Open Sampling & 91.94 & 91.08 & 89.35 & 36.92 & 75.78\\
Class Prior & 92.08 & 91.17 & 90.86 & 34.42 & 74.33 \\
BERL & 92.56 & 91.41 & 91.94 & 32.83 & 81.37\\
\hline
\textbf{COCL (Ours)} & \textbf{93.28} & \textbf{92.24} & \textbf{92.89} & \textbf{30.88} & \textbf{81.56} \\
\hline
\end{tabular}
}
\label{cifar10_table_down}
\end{subtable}
\end{table}

\begin{table}[t!]

\caption{Comparison results on CIFAR100-LT. }
\begin{subtable}{1\linewidth}
\centering
\footnotesize

\caption{\centering{Comparison of COCL to OE and OCL on six OOD datasets.
}}
\label{cifar100_table_top}
\scalebox{0.85}{
\begin{tabular}{c|c|c|c|c|c}
\hline
OOD Dataset & Method & AUC$\uparrow$ & AP-in$\uparrow$ & AP-out$\uparrow$ & FPR$\downarrow$ \\ 
\hline
\multirow{3}{*}{Texture} & OE & 76.01 & 85.28 & 57.47 & 87.45 \\ 
 & OCL & 75.92 & 82.99 & 66.48 & 70.01 \\ \cline{2-6}
 & \textbf{COCL (Ours)} & \textbf{81.99} & \textbf{88.05} & \textbf{74.38} & \textbf{59.79} \\
\hline
\multirow{3}{*}{SVHN} & OE & 81.82 & 73.25 & 89.10 & 80.98 \\
 & OCL & 78.64 & 69.21 & 86.26 & 86.38 \\ \cline{2-6}
 & \textbf{COCL (Ours)} & \textbf{89.20} & \textbf{81.57} & \textbf{94.21} & \textbf{54.46} \\
\hline
\multirow{3}{*}{CIFAR10} & OE & \textbf{62.60} & \textbf{66.16} & \textbf{57.77} & \textbf{93.53} \\
 & OCL & 60.29 & 63.21 & 55.71 & 94.22 \\ \cline{2-6}
 & \textbf{COCL (Ours)} & 62.05 & 66.14 & 56.82 & 93.88 \\
\hline
\multirow{3}{*}{\shortstack{Tiny \\ ImageNet}} & OE & 68.22 & 79.36 & 51.82 & 88.54 \\
 & OCL & 69.56 & 79.97 & 54.47 & 85.91 \\ \cline{2-6}
 & \textbf{COCL (Ours)} & \textbf{71.87} & \textbf{81.89} & \textbf{57.12} & \textbf{83.93} \\
\hline
\multirow{3}{*}{LSUN} & OE & 76.81 & 85.33 & 60.94 & 83.79 \\
 & OCL & 79.14 & 86.56 & 66.58 & 75.07 \\ \cline{2-6}
 & \textbf{COCL (Ours)} & \textbf{84.10} & \textbf{89.89} & \textbf{69.80} & \textbf{74.67} \\
\hline
\multirow{3}{*}{Place365} & OE & 75.68 & 60.99 & 86.51 & 83.55 \\
 & OCL & 77.81 & 62.80 & 88.39 & 79.97 \\ \cline{2-6}
 & \textbf{COCL (Ours)} & \textbf{80.30} & \textbf{68.65} & \textbf{89.16} & \textbf{77.83} \\
\hline
\multirow{3}{*}{Average} & OE & 73.52 & 75.06 & 67.27 & 86.30 \\
 & OCL & 73.56 & 74.12 & 69.65 & 81.93 \\ \cline{2-6}
 & \textbf{COCL (Ours)} & \textbf{78.25} & \textbf{79.37} & \textbf{73.58} & \textbf{74.09} \\
\hline
\end{tabular}
}\\

\caption{\centering{Comparison results with different competing methods. The results are averaged over the six OOD test datasets in (a).}}
\scalebox{0.85}{
\begin{tabular}{c|c|c|c|c|c}
\hline
Method & AUC$\uparrow$ & AP-in$\uparrow$ & AP-out$\uparrow$ & FPR$\downarrow$ & ACC$\uparrow$ \\ 
\hline
MSP & 63.93 & 64.71 & 60.76 & 89.71 & 40.51\\
OE & 73.52 & 75.06 & 67.27 & 86.30 & 39.42\\
EnergyOE & 76.40 & 77.32 & 72.24 & 76.33 & 41.32\\
OCL & 73.56 & 74.12 & 69.65 & 81.93 & 41.54 \\
PASCL & 73.32 & 74.84 & 67.18 & 79.38 & 43.10\\
Open Sampling & 74.37 & 75.80 & 70.42 & 78.18 & 40.87\\
Class Prior & 76.03 & 77.31 & 72.26 & 76.43 & 40.77 \\
BERL & 77.75 & 78.61 & 73.10 & 74.86 & 45.88\\
\hline
\textbf{COCL (Ours)} & \textbf{78.25} & \textbf{79.37} & \textbf{73.58} & \textbf{74.09} & \textbf{46.41} \\
\hline
\end{tabular}
}
\label{cifar100_table_down}
\end{subtable}
\end{table}

\section{Experiments}
\subsection{Experiment Settings}
\subsubsection{Datasets}
We use three popular long-tailed image classification datasets as ID data, including CIFAR10-LT \cite{cao2019learning}, CIFAR100-LT \cite{cao2019learning}, and ImageNet-LT \cite{liu2019large}. Following \cite{wang2022partial, choi2023balanced}, TinyImages 80M \cite{torralba200880} dataset is used for auxiliary OOD data to CIFAR10-LT and CIFAR100-LT,  and ImageNet-Extra \cite{wang2022partial} is used for auxiliary OOD data to ImageNet-LT. The default imbalance ratio is set to $\rho=100$ on CIFAR10-LT and
CIFAR100-LT as \cite{wang2022partial}. For OOD test set, we use six datasets CIFAR \cite{krizhevsky2009learning}, Texture \cite{cimpoi2014describing}, SVHN \cite{netzer2011reading}, LSUN \cite{yu2015lsun}, Places365 \cite{zhou2017places}, and TinyImagenet \cite{le2015tiny} introduced in the SC-OOD benchmark \cite{yang2021semantically} for the LTR task on CIFAR10-LT and  CIFAR100-LT. Following \cite{wang2022partial}, we use ImageNet-1k-OOD \cite{wang2022partial} as the OOD test set to ImageNet-LT.

\subsubsection{Evaluation Measures}
Following \cite{yang2021semantically, wang2022partial}, we use the below common metrics for OOD detection and ID classification: 
(1) FPR  is the false positive rate of OOD examples when the true positive rate of ID examples is at 95\% (as is typically done in previous OOD detection studies \cite{huang2021mos, yang2022openood, zhang2023decoupling}), this is different from the FPR in \cite{wang2022partial} that is focused on guaranteeing a 95\% true positive rate for OOD samples), 
(2) AUC computes the area under the receiver operating characteristic curve of detecting OOD samples,
(3) AP measures the area under the precision-recall curve. Depending on the selection of the positive class, AP contains AP-in which ID class samples are treated as positive, as well as AP-out where the OOD samples are regarded as positive,
and (4) ACC calculates the classification accuracy of the ID data. The reported results are averaged over six runs with different random seeds by default.

\subsubsection{Implementation Details}

We compared our approach COCL with several existing OOD detection methods on long-tailed training sets, including classical methods MSP \cite{hendrycks2016baseline}, OE \cite{hendrycks2018deep}, EnergyOE \cite{liu2020energy}, and very recently published methods PASCL \cite{wang2022partial}, Open Sampling \cite{wei2022open}, Class Prior \cite{jiang2023detecting}, and BERL \cite{choi2023balanced}. The OCL method in our results is a baseline that is trained based on Eq. \ref{OCL} only.
Following PASCL \cite{wang2022partial} and BERL \cite{choi2023balanced}, we use ResNet18 \cite{he2016deep} as our backbone on CIFAR10-LT and CIFAR100-LT, and use ResNet50 \cite{he2016deep} on ImageNet-LT.
All experiments are performed by training our model with $100$ epochs. 
The batch size is $128$ for ID data and $256$ for the auxiliary OOD data. 
More detailed implementation information is presented in Appendix B, with the full algorithm of COCL described in Appendix C.

\begin{table}[t!]
\caption{Comparison results on ImageNet-LT with ImageNet-1k-OOD as OOD test dataset.}
\centering
\footnotesize
\scalebox{0.85}{
\begin{tabular}{c|c|c|c|c|c}
\hline
Method & AUC$\uparrow$ & AP-in$\uparrow$ & AP-out$\uparrow$ & FPR$\downarrow$ & ACC$\uparrow$ \\ 
\hline
MSP & 55.78 & 35.60 & 74.18 & 94.01 & 45.36\\
OE & 68.33 & 43.87 & 82.54 & 90.98 & 44.00\\
EnergyOE & 69.43 & 45.12 & 84.75 & 76.89 & 44.42 \\
OCL & 68.67 & 43.11 & 84.15 & 77.46 & 44.77 \\
PASCL & 68.00 & 43.32 & 82.69 & 82.28 & 47.29\\
Open sampling & 69.23 & 44.21 & 84.12 & 79.37 & 45.73 \\
Class Prior & 70.43 & 45.26 & 84.82 & 77.63 & 46.83 \\
BERL & 71.16 & 45.97 & 85.63 & 76.98 & 50.42 \\
\hline
\textbf{COCL(Ours)} & \textbf{71.85} & \textbf{46.76} & \textbf{86.21} & \textbf{75.60} & \textbf{51.11} \\
\hline
\end{tabular}
}
\label{ImageNet_table}
\end{table}

\begin{table}[t!]
\caption{\centering{Comparison results on separating tail/head samples from OOD samples. The results are averaged over six OOD test datasets in the SC-OOD benchmark.}}
\begin{subtable}{1\linewidth}
\centering
\footnotesize

\caption{\centering{On separating tail samples from OOD data.}}
\label{tail_improve}
\scalebox{0.85}{
\begin{tabular}{c|ccc|ccc}
\hline
\multirow{2}{*}{Metric}  & \multicolumn{3}{c|}{CIFAR10-LT} & \multicolumn{3}{c}{CIFAR100-LT}  \\ \cline{2-7}
 & OE & OCL & COCL & OE & OCL & COCL  \\
\hline
AUC$\uparrow$ & 82.60 & 84.84 & \textbf{91.91} & 64.08 & 66.11 & \textbf{74.85} \\
AP-in$\uparrow$ & 60.47 & 61.56 & \textbf{76.98} & 34.07 & 34.97 & \textbf{47.76} \\
AP-out$\uparrow$ & 92.28 & 94.75 & \textbf{97.15} & 83.19 & 85.74 & \textbf{87.59} \\
FPR$\downarrow$ & 72.10 & 52.73 & \textbf{34.30} & 92.48 & 82.53 & \textbf{77.01} \\
\hline
\end{tabular}
}

\caption{\centering{On separating head samples from OOD data.}}
\scalebox{0.85}{
\begin{tabular}{c|ccc|ccc}
\hline
\multirow{2}{*}{Metric}  & \multicolumn{3}{c|}{CIFAR10-LT} & \multicolumn{3}{c}{CIFAR100-LT}  \\ \cline{2-7}
 & OE & OCL & COCL & OE & OCL & COCL  \\
\hline
AUC$\uparrow$ & 95.97 & 95.79 & \textbf{96.34} & 84.42 & 83.85 & \textbf{87.73} \\
AP-in$\uparrow$ & 91.09 & 88.72 & \textbf{93.34} & 70.16 & 68.44 & \textbf{73.84} \\
AP-out$\uparrow$ & 98.17 & 98.54 & \textbf{98.67} & 92.85 & 92.83 & \textbf{93.94} \\
FPR$\downarrow$ & 20.57 & 22.67 & \textbf{19.59} & 70.17 & 67.94 & \textbf{66.01} \\
\hline
\end{tabular}
}
\label{head_improve}
\end{subtable}
\end{table}

\begin{table*}[t!]
\caption{Ablation study results on CIFAR10-LT, CIFAR100-LT and ImageNet-LT. Here we also present the classification accuracy only in the tail classes of the ID dataset (ACC-t) for in-depth analysis of the performance.}
\centering
\footnotesize
\scalebox{0.85}{
\begin{tabular}{c|ccc|c|c|c|c|c|c}
\hline
ID Dataset & TCPL & DHCL & OLC & AUC$\uparrow$ & AP-in$\uparrow$ & AP-out$\uparrow$ & FPR$\downarrow$ & ACC$\uparrow$ & ACC-t$\uparrow$ \\
\hline
\hline
\multirow{7}{*}{CIFAR10-LT} & \multicolumn{3}{c|}{Baseline (OE)} & 89.76 & 89.45 & 87.22 & 53.19 & 73.59 & 55.91 \\\cline{2-10}
 & \ding{55} & \ding{55} & \ding{55} & 89.91 & 88.15 & 90.38 & 41.13 & 74.48 & 56.52 \\
 & \ding{51} & \ding{55} & \ding{55} & 91.23 & 89.47 & 91.51 & 34.27 & 74.58 & 57.10 \\
 & \ding{55} & \ding{51} & \ding{55} & 91.08 & 89.40 & 91.10 & 35.28 & 74.61 & 56.92 \\
 & \ding{55} & \ding{55} & \ding{51} & 92.06 & 91.29 & 91.78 & 34.41 & 79.40 & 76.57 \\
 & \ding{51} & \ding{51} & \ding{55} & 91.74 & 89.91 & 92.04 & 33.85 & 75.20 & 57.30 \\
 & \ding{51} & \ding{51} & \ding{51} & \textbf{93.28} & \textbf{92.24} & \textbf{92.89} & \textbf{30.88} & \textbf{81.56} & \textbf{77.90} \\
\hline
\hline
\multirow{7}{*}{CIFAR100-LT} & \multicolumn{3}{c|}{Baseline (OE)} & 73.52 & 75.06 & 67.27 & 86.30 & 39.42 & 12.59 \\\cline{2-10}
 & \ding{55} & \ding{55} & \ding{55} & 73.56 & 74.12 & 69.65 & 81.93 & 41.54 & 12.06 \\
 & \ding{51} & \ding{55} & \ding{55} & 75.14 & 75.74 & 71.25 & 78.39 & 41.93 & 13.53 \\
 & \ding{55} & \ding{51} & \ding{55} & 74.70 & 75.36 & 70.63 & 78.96 & 42.42 & 13.33 \\
 & \ding{55} & \ding{55} & \ding{51} & 75.51 & 75.83 & 71.66 & 77.57 & 45.62 & 28.44 \\
 & \ding{51} & \ding{51} & \ding{55} & 76.09 & 76.59 & 71.92 & 76.20 & 42.46 & 13.89 \\
 & \ding{51} & \ding{51} & \ding{51} & \textbf{78.25} & \textbf{79.37} & \textbf{73.58} & \textbf{74.09} & \textbf{46.41} & \textbf{29.44} \\
\hline
\hline
\multirow{7}{*}{ImageNet-LT} & \multicolumn{3}{c|}{Baseline (OE)} & 68.33 & 43.87 & 82.54 & 90.98 & 44.00 & 7.65 \\\cline{2-10}
 & \ding{55} & \ding{55} & \ding{55} & 68.67 & 43.11 & 84.15 & 77.46 & 44.77 & 8.02 \\
 & \ding{51} & \ding{55} & \ding{55} & 70.08 & 44.68 & 85.04 & 76.61 & 44.59 & 8.49 \\
 & \ding{55} & \ding{51} & \ding{55} & 69.64 & 44.11 & 84.83 & 76.62 & 45.00 & 8.43 \\
 & \ding{55} & \ding{55} & \ding{51} & 70.37 & 45.07 & 85.35 & 76.31 & 50.16 & 26.03 \\ 
 & \ding{51} & \ding{51} & \ding{55} & 70.78 & 45.19 & 85.61 & 76.26 & 45.24 & 9.92 \\
 & \ding{51} & \ding{51} & \ding{51} & \textbf{71.85} & \textbf{46.76} & \textbf{86.21} & \textbf{75.60} & \textbf{51.11} & \textbf{28.05} \\
\hline
\end{tabular}
}
\label{ablation_table}
\end{table*}

\subsection{Main Results}
Table \ref{cifar10_table_top} and Table \ref{cifar100_table_top} present 
the comparison of our COCL to the baselines OE and OCL on CIFAR10/100-LT using six commonly-used OOD test datasets. 
COCL substantially outperforms OE and OCL on both datasets across six OOD datasets except CIFAR100-LT with the CIFAR10 OOD test set where OE performs slightly better than COCL for the small semantic space. 
OCL generally achieves better performance than OE, especially in FPR, indicating that OCL can detect OOD samples better with less influence on ID classification accuracy. Our COCL improves OCL further through the three components we introduced. Note that OCL works much less effectively than OE on CIFAR100-LT vs CIFAR10 due to the difficulty of learning the outlier class given the similarity between these two datasets, which slightly drags down the performance of COCL in this case.

Table \ref{cifar10_table_down} and Table \ref{cifar100_table_down} show the comparison of COCL to state-of-the-art OOD detectors in LTR on CIFAR10-LT and CIFAR100-LT. 
COCL can improve not only OOD detection performance but also ID classification accuracy. This consistent improvement on both ID and OOD data is due to our large margin learning to reduce the biases towards head samples and OOD samples while having a customized calibration method to reinforce the LTR classification confidence.

To demonstrate the scalability of COCL, we also perform experiments on the large-scale ID dataset ImageNet-LT. The empirical results are presented in Table \ref{ImageNet_table}, 
which shows that COCL also achieves the SOTA performance in both OOD detection performance and ID classification accuracy.

\subsubsection{Performance on Separating Head/Tail Samples from OOD Samples}
To show the effectiveness of COCL in improving the capability of distinguishing OOD data from head and tail samples, we perform 
two particular OOD detection settings: one with only tail samples and OOD samples, and another one with only head samples and OOD samples. The empirical results are shown in Table \ref{tail_improve} and Table \ref{head_improve} respectively. It can be observed that (1) differentiating tail and OOD samples is often more difficult than differentiating head and OOD samples, as indicated by the AUC performance, which applies to both COCL and the two baselines, and (2) 
COCL does a better job than the two baselines in both scenarios, resulting in significantly enhanced AUC and AP values and substantially reduced FPR values.

\subsubsection{Ablation Study}
Our COCL consists of OOD-Aware Tail Class Prototype Learning (TCPL), Debiased Head Class Learning (DHCL), and Outlier-Class-Aware Logit Calibration (OLC), as elaborated in the Approach section. 
Table \ref{ablation_table} presents the results of the ablation study on these three components on all three ID datasets to show the importance of each component, with OE used as a baseline.
The method immediately below OE is another baseline OCL based on Eq. \ref{OCL}. 
The results show that (1) TCPL can largely reduce FPR, while at the same time increasing ACC-t, indicating improved performance in handling tail classes, (2) DHCL also largely reduces FPR while having similar ACC and ACC-t as OCL, indicating its effect mainly on handling head and OOD samples, (3) combining TCPL and DHCL helps leverage the strengths of both components, (4) adding the OLC component consistently improves not only the classification accuracy but also the OOD detection performance. 

\begin{figure}[t!]
  \centering
  \subfloat
  {\includegraphics[width=0.5\columnwidth]{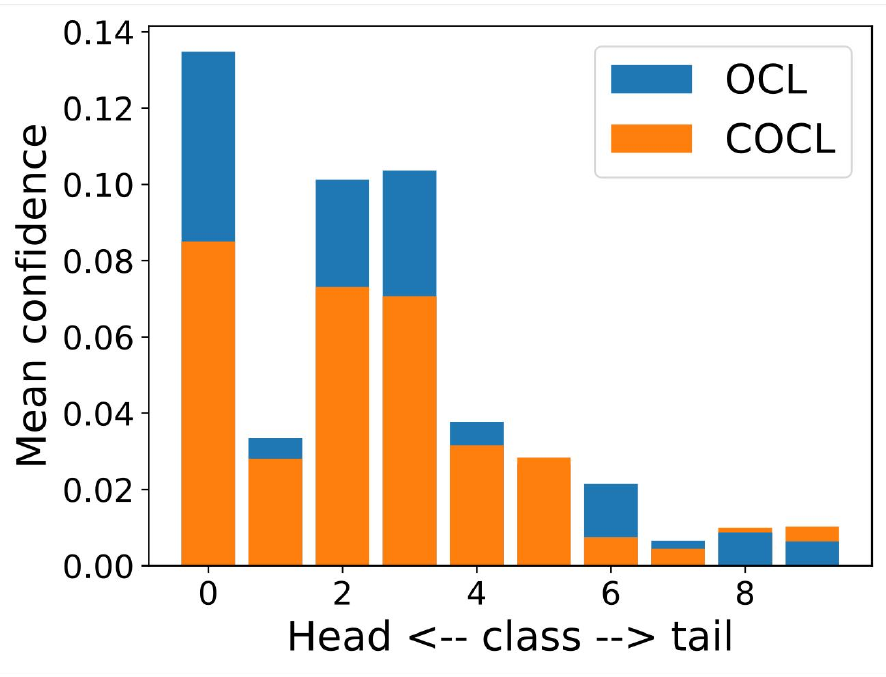}}
  \subfloat
  {\includegraphics[width=0.5\columnwidth]{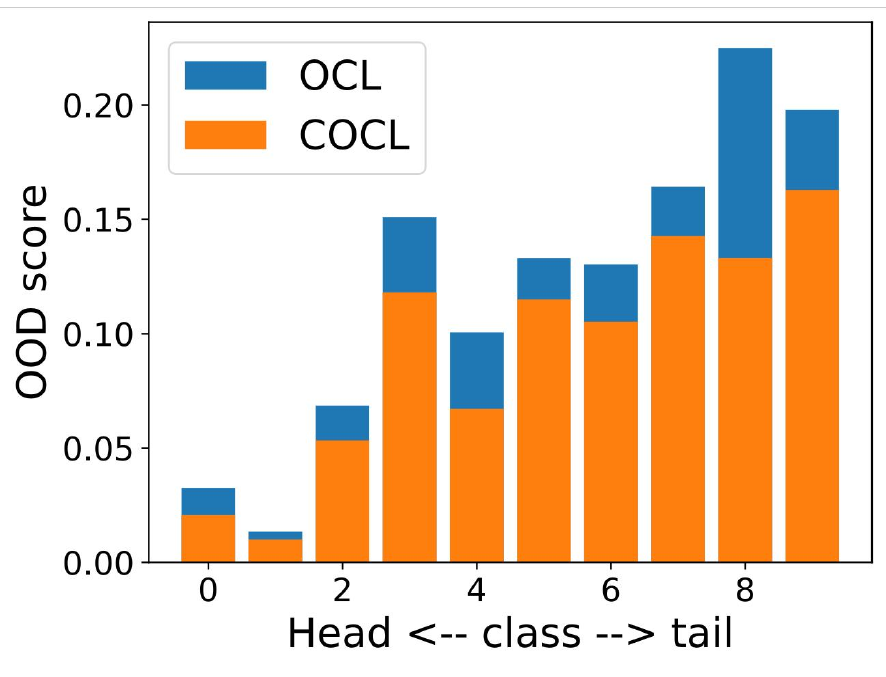}}
  \caption{Results on CIFAR10-LT. (\textbf{Left}) The mean prediction confidence of six OOD datasets belonging to each ID class. (\textbf{Right}) The mean OOD score for each ID class.}
  \label{Empirical_Analysis}
\end{figure}

\subsubsection{Qualitative Analysis}
Fig. \ref{Empirical_Analysis} presents a qualitative analysis of the prediction confidence of our method COCL on OOD samples belonging to each ID class in CIFAR10-LT (\textbf{Left}), and the mean OOD scores for each ID class (\textbf{Right}), with the results of OCL as the comparison baseline. It shows on the left panel that OCL has high confidence in predicting OOD samples as head classes, while COCL can significantly reduce these over-confident predictions. On the right panel, it is clear that our COCL can largely decrease the OOD scores for all ID class samples, particularly for tail class samples.
These results justify that the aforementioned biases towards head classes and OOD samples are effectively reduced in our model COCL, significantly enhancing COCL in distinguishing OOD samples from both head and tail classes. 
More experience are provided in Appendix D.

\section{Conclusion}
To address the OOD detection problem in LTR, we propose a novel approach, calibrated outlier class learning (COCL), to discriminate OOD samples from long-tailed ID samples. 
COCL equips the general OCL with debiased large margin learning to reduce the model biases towards head classes and OOD samples. 
It also introduces outlier-class-aware logit calibration to guarantee the long-tailed classification performance when presented with OOD samples.
Extensive experiments show that COCL significantly enhances the performance of both OOD detection and long-tailed classification on three popular LTR and OOD detection benchmarks.

\section{Acknowledgments}
W. Miao, T. Li, X. Bai, and J. Zheng were supported by National Natural Science Foundation of China (No. 62372029 and No. 62276016).
We thank the anonymous reviewers of this paper for their insightful comments that help largely enhance our work. 


\bibliography{aaai24}

\begin{thebibliography}{50}
\providecommand{\natexlab}[1]{#1}

\bibitem[{Alshammari et~al.(2022)Alshammari, Wang, Ramanan, and
  Kong}]{alshammari2022long}
Alshammari, S.; Wang, Y.-X.; Ramanan, D.; and Kong, S. 2022.
\newblock Long-tailed recognition via weight balancing.
\newblock In \emph{Proceedings of the IEEE/CVF Conference on Computer Vision
  and Pattern Recognition}, 6897--6907.

\bibitem[{Bai et~al.(2023)Bai, Liu, Wang, Hao, Feng, Chu, and
  Hu}]{bai2023effectiveness}
Bai, J.; Liu, Z.; Wang, H.; Hao, J.; Feng, Y.; Chu, H.; and Hu, H. 2023.
\newblock On the Effectiveness of Out-of-Distribution Data in Self-Supervised
  Long-Tail Learning.
\newblock \emph{arXiv preprint arXiv:2306.04934}.

\bibitem[{Cao et~al.(2019)Cao, Wei, Gaidon, Arechiga, and Ma}]{cao2019learning}
Cao, K.; Wei, C.; Gaidon, A.; Arechiga, N.; and Ma, T. 2019.
\newblock Learning imbalanced datasets with label-distribution-aware margin
  loss.
\newblock \emph{Advances in neural information processing systems}, 32.

\bibitem[{Choi, Jeong, and Choi(2023)}]{choi2023balanced}
Choi, H.; Jeong, H.; and Choi, J.~Y. 2023.
\newblock Balanced Energy Regularization Loss for Out-of-distribution
  Detection.
\newblock In \emph{Proceedings of the IEEE/CVF Conference on Computer Vision
  and Pattern Recognition}, 15691--15700.

\bibitem[{Cimpoi et~al.(2014)Cimpoi, Maji, Kokkinos, Mohamed, and
  Vedaldi}]{cimpoi2014describing}
Cimpoi, M.; Maji, S.; Kokkinos, I.; Mohamed, S.; and Vedaldi, A. 2014.
\newblock Describing textures in the wild.
\newblock In \emph{Proceedings of the IEEE conference on computer vision and
  pattern recognition}, 3606--3613.

\bibitem[{Cui et~al.(2019)Cui, Jia, Lin, Song, and Belongie}]{cui2019class}
Cui, Y.; Jia, M.; Lin, T.-Y.; Song, Y.; and Belongie, S. 2019.
\newblock Class-balanced loss based on effective number of samples.
\newblock In \emph{Proceedings of the IEEE/CVF conference on computer vision
  and pattern recognition}, 9268--9277.

\bibitem[{Gou et~al.(2023)Gou, Hu, Lv, Zhu, and Peng}]{gou2023rethinking}
Gou, Y.; Hu, P.; Lv, J.; Zhu, H.; and Peng, X. 2023.
\newblock Rethinking Image Super Resolution From Long-Tailed Distribution
  Learning Perspective.
\newblock In \emph{Proceedings of the IEEE/CVF Conference on Computer Vision
  and Pattern Recognition}, 14327--14336.

\bibitem[{He et~al.(2016)He, Zhang, Ren, and Sun}]{he2016deep}
He, K.; Zhang, X.; Ren, S.; and Sun, J. 2016.
\newblock Deep residual learning for image recognition.
\newblock In \emph{Proceedings of the IEEE conference on computer vision and
  pattern recognition}, 770--778.

\bibitem[{Hendrycks and Gimpel(2016)}]{hendrycks2016baseline}
Hendrycks, D.; and Gimpel, K. 2016.
\newblock A baseline for detecting misclassified and out-of-distribution
  examples in neural networks.
\newblock \emph{arXiv preprint arXiv:1610.02136}.

\bibitem[{Hendrycks, Mazeika, and Dietterich(2018)}]{hendrycks2018deep}
Hendrycks, D.; Mazeika, M.; and Dietterich, T. 2018.
\newblock Deep anomaly detection with outlier exposure.
\newblock \emph{arXiv preprint arXiv:1812.04606}.

\bibitem[{Hong et~al.(2023)Hong, Yao, Zhou, Zhang, and Wang}]{hong2023long}
Hong, F.; Yao, J.; Zhou, Z.; Zhang, Y.; and Wang, Y. 2023.
\newblock Long-tailed partial label learning via dynamic rebalancing.
\newblock \emph{arXiv preprint arXiv:2302.05080}.

\bibitem[{Huang and Li(2021)}]{huang2021mos}
Huang, R.; and Li, Y. 2021.
\newblock Mos: Towards scaling out-of-distribution detection for large semantic
  space.
\newblock In \emph{Proceedings of the IEEE/CVF Conference on Computer Vision
  and Pattern Recognition}, 8710--8719.

\bibitem[{Jiang et~al.(2023)Jiang, Liu, Fang, Chen, Liu, Zheng, and
  Han}]{jiang2023detecting}
Jiang, X.; Liu, F.; Fang, Z.; Chen, H.; Liu, T.; Zheng, F.; and Han, B. 2023.
\newblock Detecting Out-of-distribution Data through In-distribution Class
  Prior.
\newblock In \emph{International Conference on Machine Learning}. PMLR.

\bibitem[{Kang et~al.(2019)Kang, Xie, Rohrbach, Yan, Gordo, Feng, and
  Kalantidis}]{kang2019decoupling}
Kang, B.; Xie, S.; Rohrbach, M.; Yan, Z.; Gordo, A.; Feng, J.; and Kalantidis,
  Y. 2019.
\newblock Decoupling representation and classifier for long-tailed recognition.
\newblock \emph{arXiv preprint arXiv:1910.09217}.

\bibitem[{Kendall and Gal(2017)}]{kendall2017uncertainties}
Kendall, A.; and Gal, Y. 2017.
\newblock What uncertainties do we need in bayesian deep learning for computer
  vision?
\newblock \emph{Advances in neural information processing systems}, 30.

\bibitem[{Kingma and Ba(2014)}]{kingma2014adam}
Kingma, D.~P.; and Ba, J. 2014.
\newblock Adam: A method for stochastic optimization.
\newblock \emph{arXiv preprint arXiv:1412.6980}.

\bibitem[{Krizhevsky, Hinton et~al.(2009)}]{krizhevsky2009learning}
Krizhevsky, A.; Hinton, G.; et~al. 2009.
\newblock Learning multiple layers of features from tiny images.

\bibitem[{Krizhevsky, Sutskever, and Hinton(2017)}]{krizhevsky2017imagenet}
Krizhevsky, A.; Sutskever, I.; and Hinton, G.~E. 2017.
\newblock ImageNet classification with deep convolutional neural networks.
\newblock \emph{Communications of the ACM}, 60(6): 84--90.

\bibitem[{Le and Yang(2015)}]{le2015tiny}
Le, Y.; and Yang, X. 2015.
\newblock Tiny imagenet visual recognition challenge.
\newblock \emph{CS 231N}, 7(7): 3.

\bibitem[{Lee et~al.(2018)Lee, Lee, Lee, and Shin}]{lee2018simple}
Lee, K.; Lee, K.; Lee, H.; and Shin, J. 2018.
\newblock A simple unified framework for detecting out-of-distribution samples
  and adversarial attacks.
\newblock \emph{Advances in neural information processing systems}, 31.

\bibitem[{Leibig et~al.(2017)Leibig, Allken, Ayhan, Berens, and
  Wahl}]{leibig2017leveraging}
Leibig, C.; Allken, V.; Ayhan, M.~S.; Berens, P.; and Wahl, S. 2017.
\newblock Leveraging uncertainty information from deep neural networks for
  disease detection.
\newblock \emph{Scientific reports}, 7(1): 17816.

\bibitem[{Li et~al.(2023)Li, Chen, He, Yu, Liu, and Jia}]{li2023rethinking}
Li, J.; Chen, P.; He, Z.; Yu, S.; Liu, S.; and Jia, J. 2023.
\newblock Rethinking Out-of-distribution (OOD) Detection: Masked Image Modeling
  is All You Need.
\newblock In \emph{Proceedings of the IEEE/CVF Conference on Computer Vision
  and Pattern Recognition}, 11578--11589.

\bibitem[{Li, Cheung, and Lu(2022)}]{li2022long}
Li, M.; Cheung, Y.-m.; and Lu, Y. 2022.
\newblock Long-tailed visual recognition via gaussian clouded logit adjustment.
\newblock In \emph{Proceedings of the IEEE/CVF Conference on Computer Vision
  and Pattern Recognition}, 6929--6938.

\bibitem[{Li et~al.(2022)Li, Cao, Yuan, Fan, Yang, Feris, Indyk, and
  Katabi}]{li2022targeted}
Li, T.; Cao, P.; Yuan, Y.; Fan, L.; Yang, Y.; Feris, R.~S.; Indyk, P.; and
  Katabi, D. 2022.
\newblock Targeted supervised contrastive learning for long-tailed recognition.
\newblock In \emph{Proceedings of the IEEE/CVF Conference on Computer Vision
  and Pattern Recognition}, 6918--6928.

\bibitem[{Liu et~al.(2020)Liu, Wang, Owens, and Li}]{liu2020energy}
Liu, W.; Wang, X.; Owens, J.; and Li, Y. 2020.
\newblock Energy-based out-of-distribution detection.
\newblock \emph{Advances in neural information processing systems}, 33:
  21464--21475.

\bibitem[{Liu et~al.(2023)Liu, Ding, Tian, Pang, Belagiannis, Reid, and
  Carneiro}]{liu2023residual}
Liu, Y.; Ding, C.; Tian, Y.; Pang, G.; Belagiannis, V.; Reid, I.; and Carneiro,
  G. 2023.
\newblock Residual pattern learning for pixel-wise out-of-distribution
  detection in semantic segmentation.
\newblock In \emph{Proceedings of the IEEE/CVF International Conference on
  Computer Vision}, 1151--1161.

\bibitem[{Liu et~al.(2019)Liu, Miao, Zhan, Wang, Gong, and Yu}]{liu2019large}
Liu, Z.; Miao, Z.; Zhan, X.; Wang, J.; Gong, B.; and Yu, S.~X. 2019.
\newblock Large-scale long-tailed recognition in an open world.
\newblock In \emph{Proceedings of the IEEE/CVF conference on computer vision
  and pattern recognition}, 2537--2546.

\bibitem[{Menon et~al.(2020)Menon, Jayasumana, Rawat, Jain, Veit, and
  Kumar}]{menon2020long}
Menon, A.~K.; Jayasumana, S.; Rawat, A.~S.; Jain, H.; Veit, A.; and Kumar, S.
  2020.
\newblock Long-tail learning via logit adjustment.
\newblock \emph{arXiv preprint arXiv:2007.07314}.

\bibitem[{Nam, Jang, and Lee(2023)}]{nam2023decoupled}
Nam, G.; Jang, S.; and Lee, J. 2023.
\newblock Decoupled Training for Long-Tailed Classification With Stochastic
  Representations.
\newblock \emph{arXiv preprint arXiv:2304.09426}.

\bibitem[{Netzer et~al.(2011)Netzer, Wang, Coates, Bissacco, Wu, and
  Ng}]{netzer2011reading}
Netzer, Y.; Wang, T.; Coates, A.; Bissacco, A.; Wu, B.; and Ng, A.~Y. 2011.
\newblock Reading digits in natural images with unsupervised feature learning.

\bibitem[{Russakovsky et~al.(2015)Russakovsky, Deng, Su, Krause, Satheesh, Ma,
  Huang, Karpathy, Khosla, Bernstein et~al.}]{russakovsky2015imagenet}
Russakovsky, O.; Deng, J.; Su, H.; Krause, J.; Satheesh, S.; Ma, S.; Huang, Z.;
  Karpathy, A.; Khosla, A.; Bernstein, M.; et~al. 2015.
\newblock Imagenet large scale visual recognition challenge.
\newblock \emph{International journal of computer vision}, 115: 211--252.

\bibitem[{Sastry and Oore(2020)}]{sastry2020detecting}
Sastry, C.~S.; and Oore, S. 2020.
\newblock Detecting out-of-distribution examples with gram matrices.
\newblock In \emph{International Conference on Machine Learning}, 8491--8501.
  PMLR.

\bibitem[{Sun, Guo, and Li(2021)}]{sun2021react}
Sun, Y.; Guo, C.; and Li, Y. 2021.
\newblock React: Out-of-distribution detection with rectified activations.
\newblock \emph{Advances in Neural Information Processing Systems}, 34:
  144--157.

\bibitem[{Tan et~al.(2020)Tan, Wang, Li, Li, Ouyang, Yin, and
  Yan}]{tan2020equalization}
Tan, J.; Wang, C.; Li, B.; Li, Q.; Ouyang, W.; Yin, C.; and Yan, J. 2020.
\newblock Equalization loss for long-tailed object recognition.
\newblock In \emph{Proceedings of the IEEE/CVF conference on computer vision
  and pattern recognition}, 11662--11671.

\bibitem[{Tang et~al.(2022)Tang, Tao, Qi, Liu, and Zhang}]{tang2022invariant}
Tang, K.; Tao, M.; Qi, J.; Liu, Z.; and Zhang, H. 2022.
\newblock Invariant feature learning for generalized long-tailed
  classification.
\newblock In \emph{European Conference on Computer Vision}, 709--726. Springer.

\bibitem[{Tian et~al.(2022)Tian, Liu, Pang, Liu, Chen, and
  Carneiro}]{tian2022pixel}
Tian, Y.; Liu, Y.; Pang, G.; Liu, F.; Chen, Y.; and Carneiro, G. 2022.
\newblock Pixel-wise energy-biased abstention learning for anomaly segmentation
  on complex urban driving scenes.
\newblock In \emph{European Conference on Computer Vision}, 246--263. Springer.

\bibitem[{Torralba, Fergus, and Freeman(2008)}]{torralba200880}
Torralba, A.; Fergus, R.; and Freeman, W.~T. 2008.
\newblock 80 million tiny images: A large data set for nonparametric object and
  scene recognition.
\newblock \emph{IEEE transactions on pattern analysis and machine
  intelligence}, 30(11): 1958--1970.

\bibitem[{Wang et~al.(2022)Wang, Zhang, Zhu, Zheng, Li, Smola, and
  Wang}]{wang2022partial}
Wang, H.; Zhang, A.; Zhu, Y.; Zheng, S.; Li, M.; Smola, A.~J.; and Wang, Z.
  2022.
\newblock Partial and asymmetric contrastive learning for out-of-distribution
  detection in long-tailed recognition.
\newblock In \emph{International Conference on Machine Learning}, 23446--23458.
  PMLR.

\bibitem[{Wang et~al.(2020{\natexlab{a}})Wang, Li, Kang, Li, Liew, Tang, Hoi,
  and Feng}]{wang2020devil}
Wang, T.; Li, Y.; Kang, B.; Li, J.; Liew, J.; Tang, S.; Hoi, S.; and Feng, J.
  2020{\natexlab{a}}.
\newblock The devil is in classification: A simple framework for long-tail
  instance segmentation.
\newblock In \emph{Computer Vision--ECCV 2020: 16th European Conference,
  Glasgow, UK, August 23--28, 2020, Proceedings, Part XIV 16}, 728--744.
  Springer.

\bibitem[{Wang et~al.(2020{\natexlab{b}})Wang, Lian, Miao, Liu, and
  Yu}]{wang2020long}
Wang, X.; Lian, L.; Miao, Z.; Liu, Z.; and Yu, S.~X. 2020{\natexlab{b}}.
\newblock Long-tailed recognition by routing diverse distribution-aware
  experts.
\newblock \emph{arXiv preprint arXiv:2010.01809}.

\bibitem[{Wang et~al.(2023)Wang, Li, Chen, Lim, Torralba, Zhao, and
  Wang}]{wang2023detecting}
Wang, Z.; Li, Y.; Chen, X.; Lim, S.-N.; Torralba, A.; Zhao, H.; and Wang, S.
  2023.
\newblock Detecting everything in the open world: Towards universal object
  detection.
\newblock In \emph{Proceedings of the IEEE/CVF Conference on Computer Vision
  and Pattern Recognition}, 11433--11443.

\bibitem[{Wei et~al.(2022{\natexlab{a}})Wei, Tao, Xie, Feng, and
  An}]{wei2022open}
Wei, H.; Tao, L.; Xie, R.; Feng, L.; and An, B. 2022{\natexlab{a}}.
\newblock Open-sampling: Exploring out-of-distribution data for re-balancing
  long-tailed datasets.
\newblock In \emph{International Conference on Machine Learning}, 23615--23630.
  PMLR.

\bibitem[{Wei et~al.(2022{\natexlab{b}})Wei, Xie, Cheng, Feng, An, and
  Li}]{wei2022mitigating}
Wei, H.; Xie, R.; Cheng, H.; Feng, L.; An, B.; and Li, Y. 2022{\natexlab{b}}.
\newblock Mitigating neural network overconfidence with logit normalization.
\newblock In \emph{International Conference on Machine Learning}, 23631--23644.
  PMLR.

\bibitem[{Yang et~al.(2021)Yang, Wang, Feng, Yan, Zheng, Zhang, and
  Liu}]{yang2021semantically}
Yang, J.; Wang, H.; Feng, L.; Yan, X.; Zheng, H.; Zhang, W.; and Liu, Z. 2021.
\newblock Semantically coherent out-of-distribution detection.
\newblock In \emph{Proceedings of the IEEE/CVF International Conference on
  Computer Vision}, 8301--8309.

\bibitem[{Yang et~al.(2022)Yang, Wang, Zou, Zhou, Ding, Peng, Wang, Chen, Li,
  Sun et~al.}]{yang2022openood}
Yang, J.; Wang, P.; Zou, D.; Zhou, Z.; Ding, K.; Peng, W.; Wang, H.; Chen, G.;
  Li, B.; Sun, Y.; et~al. 2022.
\newblock Openood: Benchmarking generalized out-of-distribution detection.
\newblock \emph{Advances in Neural Information Processing Systems}, 35:
  32598--32611.

\bibitem[{Yu et~al.(2015)Yu, Seff, Zhang, Song, Funkhouser, and
  Xiao}]{yu2015lsun}
Yu, F.; Seff, A.; Zhang, Y.; Song, S.; Funkhouser, T.; and Xiao, J. 2015.
\newblock Lsun: Construction of a large-scale image dataset using deep learning
  with humans in the loop.
\newblock \emph{arXiv preprint arXiv:1506.03365}.

\bibitem[{Yu et~al.(2023)Yu, Shin, Lee, Jun, and Lee}]{yu2023block}
Yu, Y.; Shin, S.; Lee, S.; Jun, C.; and Lee, K. 2023.
\newblock Block Selection Method for Using Feature Norm in Out-of-Distribution
  Detection.
\newblock In \emph{Proceedings of the IEEE/CVF Conference on Computer Vision
  and Pattern Recognition}, 15701--15711.

\bibitem[{Zhang and Xiang(2023)}]{zhang2023decoupling}
Zhang, Z.; and Xiang, X. 2023.
\newblock Decoupling MaxLogit for Out-of-Distribution Detection.
\newblock In \emph{Proceedings of the IEEE/CVF Conference on Computer Vision
  and Pattern Recognition}, 3388--3397.

\bibitem[{Zhou et~al.(2017)Zhou, Lapedriza, Khosla, Oliva, and
  Torralba}]{zhou2017places}
Zhou, B.; Lapedriza, A.; Khosla, A.; Oliva, A.; and Torralba, A. 2017.
\newblock Places: A 10 million image database for scene recognition.
\newblock \emph{IEEE transactions on pattern analysis and machine
  intelligence}, 40(6): 1452--1464.

\bibitem[{Zhu et~al.(2023)Zhu, Cheng, Zhang, and Liu}]{zhu2023openmix}
Zhu, F.; Cheng, Z.; Zhang, X.-Y.; and Liu, C.-L. 2023.
\newblock OpenMix: Exploring Outlier Samples for Misclassification Detection.
\newblock In \emph{Proceedings of the IEEE/CVF Conference on Computer Vision
  and Pattern Recognition}, 12074--12083.

\end{thebibliography}

\newpage

\section{A. More Discussion about \\ Outlier-Class-Aware Logit Calibration}

Outlier-class-aware logit calibration not only improves long-tailed classification confidence but also enhances OOD detection performance. We provide more theory discussion about outlier-class-aware logit calibration as follows. Formally, since an input $x$ is inferred based on Eq. 8 in the Approach section, we use the posterior probability of the outlier class to derive for $i$-th ID class where $i$ is one of the ID classes.
\begin{equation}
\frac{\partial P(y=k+1|x)}{\partial f_i(x)} = - \frac{1}{n_i} \frac{e^{f_{K+1}(x)} \cdot e^{f_i(x)}}{(\sum_{j=1}^{k+1}e^{f_j(x) - \log n_j})^2}. \label{E8} 
\end{equation}
The first-order partial derivative of the outlier class in each ID class is lower than zero.
Thus, the posterior probability after outlier-class-aware logit calibration is inverse to each ID class confidence. 
That is, the higher the ID class confidence, the lower the outlier class scores.
Moreover, the different prior probability $n$ means that the different roles for each class to the outlier class score. Specifically, consider a tail class $i$ and a head class $j$, we have
\begin{align}
\begin{aligned}
& \left| \frac{\partial P(y=k+1|x)}{\partial f_i(x)} \right| - \left| \frac{\partial P(y=k+1|x)}{\partial f_j(x)} \right| \\
& = (\frac{e^{f_i(x)}}{n_i} - \frac{e^{f_j(x)}}{n_j}) \frac{e^{f_{k+1}(x)}}{(\sum_{j=1}^{k+1}e^{f_j(x) - \log n_j})^2}. \label{E9} 
\end{aligned}
\end{align}
When an input sample $x$ has the same confidence in the classes $i$ and $j$ that $f_i(x) = f_j(x)$, then the more tailed class has smaller prior probability $n_i < n_j$ which leads to the higher partial derivative for the outlier class score in class $i$ than class $j$ that $\left| \frac{\partial P(y=K+1|x)}{\partial f_i(x)} \right| > \left| \frac{\partial P(y=K+1|x)}{\partial f_j(x)} \right|$. 

This indicates that the tail class confidence plays a more important role than the head class in decreasing outlier class scores. 
Particularly, although tail samples always have high OOD scores that are easily mistaken for OOD samples, these tail samples also attain higher ID prediction confidence within their respective tail classes compared to OOD samples. Relatively, OOD samples tend to possess high confidence in head classes that easily confuse with head samples, but their confidence levels in tail classes remain lower than those of tail samples. Therefore, with outlier-class-aware logit calibration during the inference phase, OOD samples can be more effectively distinguished from tail samples due to the significant decrease in outlier class scores brought about by the higher tail confidence in tail classes.
And head classes are already detected well with OOD data, so outlier-class-aware logit calibration rarely harms head classes.

\section{B. Implementation Details}
For experiments on CIFAR10-LT and CIFAR100-LT, we train our model based on ResNet18 via Algorithm \ref{alg:algorithm} for 100 epochs using Adam \cite{kingma2014adam} optimizer with initial learning rate $1 \times 10^{-3}$. We decay the learning rate to zero using a cosine annealing learning rate schedule. Our auxiliary dataset is a subset of TinyImages80M with 300K images, following PASCAL \cite{wang2022partial}. For ImageNet-LT, we train our model based on ResNet50 via Algorithm \ref{alg:algorithm} for $100$ epochs using SGD optimizer with an initial learning rate $0.1$. We also decay the learning rate to zero using a cosine annealing learning rate scheduler. Our auxiliary dataset is a subset of ImageNet22k, following PASCAL. 
Empirically, we set  $\gamma = 0.05$, $\alpha = 0.05$, $\beta = 0.1$, $t=0.07$, $\tau = 1$, $margin = 1$.

\section{C. The COCL Algorithm}
\begin{algorithm}[ht]
\caption{COCL}
\label{alg:algorithm}
\textbf{Input}: training dataset $ \mathcal {D}_{in}^{train} $; auxiliary dataset $ \mathcal {D}_{out}^{train} $; \\ unlabeled dataset $ \mathcal {D}_{in \cup out}^{test} $ \\
\textbf{Training}
\begin{algorithmic}[1] 
\FOR{each iteration} 
\STATE Sample a mini-batch of ID training data, \\
$ \left\{ (x_i^{in}, y_i) \right\}_{i=1}^n $ from $ \mathcal {D}_{in}^{train} $
\STATE Sample a mini-batch of OOD auxiliary data, \\
$ \left\{ (x_i^{out}) \right\}_{i=1}^n $ from $ \mathcal {D}_{out}^{train} $
\STATE Perform common gradient descent on model $f$ with $\mathcal{L}_{total}$ based on Eq. 7
\ENDFOR
\end{algorithmic}
\textbf{Inference}
\begin{algorithmic}[1] 
\FOR{each sample $x$ in dataset $ \mathcal {D}_{in \cup out}^{test} $} 
\STATE inference $x$ on model $f$ base on Eq. 8
\ENDFOR
\end{algorithmic}
\end{algorithm}

\begin{table*}[t!]
\caption{Comparison results of imbalance ratio between COCL and OCL on CIFAR10-LT. The results are averaged over six OOD test datasets in the SC-OOD benchmark.}
\centering
\begin{tabular}{c|c|c|c|c|c|c}
\hline
Imbalance Ratio& Method & AUC$\uparrow$ & AP-in$\uparrow$ & AP-out$\uparrow$ & FPR$\downarrow$ & ACC$\uparrow$ \\ 
\hline
\multirow{2}{*}{$\rho=100$} & OCL & 89.91 & 88.15 & 90.38 & 41.13 & 74.48\\
& \textbf{COCL(Ours)} & \textbf{93.28} & \textbf{92.24} & \textbf{92.89} & \textbf{30.88} & \textbf{81.56} \\
\hline
\multirow{2}{*}{$\rho=50$} & OCL & 92.89 & 91.97 & 92.78 & 31.43 & 80.13\\
& \textbf{COCL(Ours)} & \textbf{94.30} & \textbf{93.85} & \textbf{93.31} & \textbf{26.98} & \textbf{84.89} \\
\hline
\multirow{2}{*}{$\rho=10$} & OCL & 95.20 & 94.74 & 95.02 & 23.31 & 88.36\\
& \textbf{COCL(Ours)} & \textbf{95.71} & \textbf{95.12} & \textbf{95.33} & \textbf{20.91} & \textbf{89.65} \\
\hline
\end{tabular}
\label{Imbalance_ratio}
\end{table*}

\begin{table*}[t!]
\caption{Comparison results of model structure between COCL and OCL on CIFAR10-LT. The results are averaged over six OOD test datasets in the SC-OOD benchmark.}
\centering
\begin{tabular}{c|c|c|c|c|c|c}
\hline
Model& Method & AUC$\uparrow$ & AP-in$\uparrow$ & AP-out$\uparrow$ & FPR$\downarrow$ & ACC$\uparrow$ \\ 
\hline
\multirow{2}{*}{ResNet18} & OCL & 89.91 & 88.15 & 90.38 & 41.13 & 74.48\\
& \textbf{COCL(Ours)} & \textbf{93.28} & \textbf{92.24} & \textbf{92.89} & \textbf{30.88} & \textbf{81.56} \\
\hline
\multirow{2}{*}{ResNet34} & OCL & 91.30 & 89.68 & 91.53 & 37.03 & 74.98\\
& \textbf{COCL(Ours)} & \textbf{93.52} & \textbf{92.93} & \textbf{92.83} & \textbf{30.74} & \textbf{81.75} \\
\hline
\end{tabular}
\label{Model_structures_cifar10}
\end{table*}

\section{D. More Experimental Results}
\subsection{D.1 More Ablation Study}
\subsubsection{Imbalance Ratio $\rho$}
In the Experiments section, we use an imbalance ratio $\rho=100$ on both CIFAR10-LT and CIFAR100-LT. In this section, we show that our method can work well under different imbalance ratios. Specifically, we conduct experiments on CIFAR10-LT with $\rho=50$ and $\rho=10$. The results are shown in Table \ref{Imbalance_ratio}. Our method also outperforms the OCL baseline by a considerable margin. Furthermore, our approach COCL performs better in a more imbalanced dataset.

\subsubsection{Model Structures}
In the Experiments section, we use the standard ResNet18 as the backbone model on both CIFAR10-LT and CIFAR100-LT. In this section, we show that our method can work well under different model structures. So we conduct experiments using the standard ResNet34. The results are shown in Table \ref{Model_structures_cifar10}. Our method also outperforms the OCL baseline by a considerable margin on ResNet34. Furthermore, our method can both obtain slightly enhancement on OOD detection and ID classification without any hyperparameter adjustment in ResNet34.

\subsubsection{Ablation Study on Temperature $t$}
The results are shown in Table \ref{Ablation_t}. $t$ is a hyperparameter in OOO-aware tail class prototype learning. The performance of our method is stable with respect to different $t$ values.

\begin{table}[ht]
\caption{Comparison results of different temperature $t$ of our approach COCL on CIFAR10-LT. The results are averaged over six OOD test datasets in the SC-OOD benchmark}
\centering
\begin{tabular}{c|c|c|c|c|c}
\hline
 $t$ & AUC$\uparrow$ & AP-in$\uparrow$ & AP-out$\uparrow$ & FPR$\downarrow$ & ACC$\uparrow$ \\ 
\hline
 0.02 & 92.22 & 90.96 & 91.57 & 31.76 & 80.77 \\
 0.05 & 93.12 & 92.14 & 92.25 & \textbf{30.26} & 81.07 \\
\textbf{0.07} & \textbf{93.28} & \textbf{92.24} & \textbf{92.89} & 30.88 & \textbf{81.56}\\
0.1 & 92.95 & 92.09 & 91.54 & 31.27 & 78.70 \\
0.2 & 92.39 & 91.76 & 90.82 & 35.67 & 77.87 \\
\hline
\end{tabular}
\label{Ablation_t}
\end{table}

\subsubsection{Ablation Study on $margin$}
The results are shown in Table \ref{Ablation_mar}. $margin$ is a user-defined hyperparameter that specifies the margin between the one-class OOD description region and the head samples. if the margin is too small, we cannot effectively distinguish OOD data and head classes, but if the margin is too large, it may compass the feature region of ID classes, leading to lower ID classification accuracy.

\begin{table}[ht]
\caption{Comparison results of different $margin$ of our approach COCL on CIFAR10-LT. The results are averaged over six OOD test datasets in the SC-OOD benchmark}
\centering
\begin{tabular}{c|c|c|c|c|c}
\hline
 $margin$ & AUC$\uparrow$ & AP-in$\uparrow$ & AP-out$\uparrow$ & FPR$\downarrow$ & ACC$\uparrow$ \\ 
 \hline
 0.5 & 93.06 & \textbf{92.51} & 91.70 & 32.37 & 81.50 \\
\textbf{1} & \textbf{93.28} & 92.24 & \textbf{92.89} & \textbf{30.88} & \textbf{81.56} \\
 2 & 93.25 & 92.10 & 92.22 & 30.90 & 81.15 \\
\hline
\end{tabular}
\label{Ablation_mar}
\end{table}

\subsubsection{Ablation Study on Hyperparameter $\tau$}
The results are shown in Table \ref{Ablation_tau}. $\tau$ is a hyperparameter to balance how much we want to bring in the prior of the outlier class. Usually, we set $\tau=1$ which is fit to the training sample size prior. Moreover, the performance of our method is stable with respect to different $\tau$ values.

\begin{table}[ht]
\caption{Comparison results of hyperparameter $\tau$ with our approach COCL on CIFAR10-LT. The results are averaged over six OOD test datasets in the SC-OOD benchmark}
\centering
\begin{tabular}{c|c|c|c|c|c}
\hline
 $\tau$ & AUC$\uparrow$ & AP-in$\uparrow$ & AP-out$\uparrow$ & FPR$\downarrow$ & ACC$\uparrow$ \\ 
\hline
 0.5 & 92.98 & 92.29 & 91.62 & 32.40 & 80.98 \\
\textbf{1} & \textbf{93.28} & 92.24 & \textbf{92.89} & \textbf{30.88} & \textbf{81.56} \\
 2 & 93.05 & \textbf{92.85} & 91.40 & 32.49 & 80.88 \\
\hline
\end{tabular}
\label{Ablation_tau}
\end{table}

\begin{table*}[t!]
\caption{Average AUROC on CIFAR10-LT using ResNet18 depending on hyperparameter $\gamma$, $\alpha$, and $\beta$. The results are averaged over six OOD test datasets in the SC-OOD benchmark}
\centering
\begin{tabular}{|c|c|c|c|c|c|c|c|c|c|c|}
\hline
\multicolumn{11}{|c|}{Average AUROC} \\ 
\hline
\multirow{4}{*}{Hyperparameter $\gamma$} & \multicolumn{10}{c|}{Hyperparameter $\alpha$} \\ 
\cline{2-11}
 & \multicolumn{2}{c|}{0.01} & \multicolumn{3}{c|}{0.02} & \multicolumn{3}{c|}{0.05} & \multicolumn{2}{c|}{0.1}  \\ 
 \cline{2-11}
 & \multicolumn{10}{c|}{Hyperparameter $\beta$} \\ 
\cline{2-11}
 & 0.01 & 0.02 & 0.01 & 0.02 & 0.05 & 0.02 & 0.05 & 0.1 & 0.05 & 0.1\\ 
\hline
0.01  & 91.22 & 91.81 & 91.45 & 91.97 & 92.38 & 91.79 & 92.19 & 91.79 & 91.90 & 92.47\\ 
\hline
0.02  & 92.07 & 92.39 & 92.34 & 92.41 & 92.72 & 92.71 & 92.75 & 92.71 & 92.77 & 92.70\\ 
\hline
0.05  & 93.04 & 92.96 & 92.94 & 92.98 & 93.13 & 92.75 & 93.17 & \textbf{93.28} & 93.18 & 92.75\\ 
\hline
0.1  & 93.19 & 92.86 & 92.99 & 92.94 & 93.03 & 93.13 & 93.14 & 93.13 & 93.20 & 93.23\\ 
\hline
\end{tabular}
\label{hyper_auc}
\end{table*}

\begin{table*}[t!]
\caption{Accuracy on CIFAR10-LT using ResNet18 depending on hyperparameter $\gamma$, $\alpha$, and $\beta$. The results are averaged over six OOD test datasets in the SC-OOD benchmark}
\centering
\begin{tabular}{|c|c|c|c|c|c|c|c|c|c|c|}
\hline
\multicolumn{11}{|c|}{Accuracy (ACC)} \\ 
\hline
\multirow{4}{*}{Hyperparameter $\gamma$} & \multicolumn{10}{c|}{Hyperparameter $\alpha$} \\ 
\cline{2-11}
 & \multicolumn{2}{c|}{0.01} & \multicolumn{3}{c|}{0.02} & \multicolumn{3}{c|}{0.05} & \multicolumn{2}{c|}{0.1}  \\ 
 \cline{2-11}
 & \multicolumn{10}{c|}{Hyperparameter $\beta$} \\ 
\cline{2-11}
 & 0.01 & 0.02 & 0.01 & 0.02 & 0.05 & 0.02 & 0.05 & 0.1 & 0.05 & 0.1\\ 
\hline
0.01  & 79.61 & 79.58 & 79.17 & 80.50 & 80.86 & 79.37 & 80.05 & 79.37 & 79.58 & 80.82\\ 
\hline
0.02  & 79.30 & 80.02 & 79.47 & 80.33 & 79.93 & 79.14 & 80.13 & 79.14 & 80.38 & 80.09\\ 
\hline
0.05  & 80.68 & 80.36 & 80.54 & 80.71 & 80.93 & 80.31 & 80.97 & \textbf{81.56} & 80.73 & 80.87 \\ 
\hline
0.1  & 80.30 & 80.17 & 80.44 & 80.10 & 80.25 & 80.77 & 80.03 & 80.77 & 80.04 & 80.39 \\ 
\hline
\end{tabular}
\label{hyper_ACC}
\end{table*}

\subsubsection{Ablation Study on Hyperparameter $\gamma$, $\alpha$, and $\beta$}
As discussed in the Approach section, we introduce the debiased large margin learning in the outlier class learning during training. More specifically, hyperparameter $\gamma$ is set to balance the ID class learning and outlier class learning. Thus, the higher the $\gamma$, the LTR models are more preferred to the outliers. With the hyperparameter $\gamma$ improved, it has a better OOD detection performance but could be harmful to the ID classification. Conversely, if we set $\gamma=0$ means that we do not use the outlier class learning. As seen in Table \ref{hyper_auc} and Table \ref{hyper_ACC}, AUROC and accuracy also show similar tendencies as discussed above. So it needs to choose a suitable $\gamma$ corresponding to the OOD training batch size. Furthermore, hyperparameters $\alpha$ and $\beta$ are set to balance the model biases towards OOD samples and head samples, and having hyperparameter $\beta$ slightly larger than hyperparameter $\alpha$ is usually beneficial to debiased large margin learning as shown in Table \ref{hyper_auc} and Table \ref{hyper_ACC}. Moreover, Hyperparameter $\gamma$ is sensitivity to $\alpha$ and $\beta$. Finally, $\gamma=0.05$ with corresponding $\alpha=0.05$ and $\beta=0.1$ is generally recommended and used by default in our experiments. 

\begin{table}[t!]
\caption{Comparison results of the different percentage of head/tail classes used on CIFAR10-LT. The results are averaged over six OOD test datasets in the SC-OOD benchmark}
\centering
\begin{tabular}{c|c|c|c|c|c}
\hline
$k$ (tail) & AUC$\uparrow$ & AP-in$\uparrow$ & AP-out$\uparrow$ & FPR$\downarrow$ & ACC$\uparrow$ \\ 
\hline
50\% & 93.01 & 92.41 & 91.77 & 31.26 & 81.33 \\
\textbf{40\%} & \textbf{93.28} & 92.24 & \textbf{92.89} & \textbf{30.88} & \textbf{81.56} \\
30\% & 93.21 & \textbf{92.70} & 92.29 & 31.17 & 81.48 \\
\hline
0\% & 92.27 & 91.00 & 91.78 & 32.64 & 79.37 \\
\hline
\hline
$p$ (head) & AUC$\uparrow$ & AP-in$\uparrow$ & AP-out$\uparrow$ & FPR$\downarrow$ & ACC$\uparrow$ \\ 
\hline
50\% & 92.98 & 92.09 & 92.18 & 31.96 & 80.37 \\
\textbf{40\%} & \textbf{93.28} & 92.24 & \textbf{92.89} & \textbf{30.88} & \textbf{81.56} \\
30\% & 93.25 & \textbf{92.46} & 92.43 & 32.01 & 81.09 \\
\hline
0\% & 92.72 & 91.73 & 91.68 & 32.97 & 78.65 \\
\hline
\end{tabular}
\label{percentage}
\end{table}

\begin{table}[t!]
\caption{Comparison results on Synthetic OOD dataset with CIFAR10-LT.}
\centering
\footnotesize
\begin{tabular}{c|c|c|c|c|c}
\hline
Dataset & Method & AUC$\uparrow$ & AP-in$\uparrow$ & AP-out$\uparrow$ & FPR$\downarrow$ \\ 
\hline
\multirow{5}{*}{Gaussian} & OE & 97.15 & 99.28 & 92.19 & 5.11\\
& OS & 99.55 & 95.72 & 86.37 & 1.20\\
& BERL & \textbf{99.76} & 99.34 & 99.16 & 0.49\\ \cline{2-6}
& OCL & 99.45 & 99.03 & 98.64 & 0.1 \\ 
& \textbf{COCL} & 99.68 & \textbf{99.79} & \textbf{99.39} & \textbf{0.02} \\
\hline
\multirow{5}{*}{Rademacher} & OE & 99.12 & 90.36 & 98.24 & 1.57\\
& OS & 97.90 & 78.00 & 82.44 & 3.25\\
& BERL & 99.00 & 99.06 & 96.26 & 1.42 \\ \cline{2-6}
& OCL & 99.18 & 99.37 & 98.70 & 1.73 \\ 
& \textbf{COCL} & \textbf{99.76} & \textbf{99.84} & \textbf{99.56} & \textbf{0.01} \\
\hline
\multirow{5}{*}{Blobs} & OE & 43.80 & 54.31 & 43.12 & 84.15\\
& OS & 91.50 & 66.51 & 76.31 & 33.65 \\
& BERL & 93.18 & 96.87 & 89.34 & 22.20\\ \cline{2-6}
& OCL & 96.87 & 98.69 & 96.79 & 7.04 \\ 
& \textbf{COCL} & \textbf{98.75} & \textbf{99.17} & \textbf{97.49} & \textbf{1.04} \\
\hline
\multirow{5}{*}{Average} & OE & 80.08 & 81.32 & 77.85 & 30.27 \\
& OS & 96.32 & 80.08 & 81.71 & 12.7 \\
& BERL & 97.32 & 98.42 & 94.92 & 9.04 \\ \cline{2-6}
& OCL & 98.50 & 99.03 & 98.04 & 2.96\\ 
& \textbf{COCL} & \textbf{99.40} & \textbf{99.60} & \textbf{98.81} & \textbf{0.35} \\
\hline
\end{tabular}
\label{synthetic}
\end{table}

\subsubsection{Ablation Study on the Percentage of Head/Tail Classes to Use}
As mentioned in the Approach section, we only apply OOD-aware tail class prototype learning between tail-class samples and OOD data to mitigate the model bias towards OOD samples. Moreover, we only apply debiased head class learning on head-class samples and OOD data to mitigate the model bias towards head samples. A key hyperparameter here is the threshold to define the separation of head and tail classes: the $k$ (in percentage) classes with the least training samples are defined as tail classes and the $p$ (in percentage) classes with the most training samples are defined as head classes. 
Ablation study on $k$ and $p$ is provided in Table \ref{percentage}. As we can see, on CIFAR10-LT, the best results are achieved around $k = 40\%$ and $p = 40\%$ (the default value in our experiments). The results at $k = 40\%$ are considerably better than those at $k$ = 0\% (without OOD-aware tail class prototype learning) and the results at $p = 40\%$ are considerably better than those at $p$ = 0\% (without debiased head class learning), showing the importance of our debiased large margin learning design in the outlier class learning. Also, the performance of our method is stable with respect to $k$ and $p$ within a large range (e.g. $k \in [30\%, 50\%]$, $p \in [30\%, 50\%]$).

\subsection{D.2 Experiment Results on Synthetic OOD Datasets}
To demonstrate the superiority of our approach COCL, we comprehensively evaluate OOD detectors on synthetic data using ResNet18 with CIFAR10-LT, following OS \cite{wei2022open} and BERL \cite{choi2023balanced}, including Gaussian, Rademacher, and Blobs. Specifically, \textit{Gaussian} noises have each dimension i.i.d. sampled from an isotropic Gaussian distribution. \textit{Rademacher} noises are images where each dimension is $\textnormal{-} 1$ or $1$ with equal probability, so each dimension is sampled from a symmetric Rademacher distribution. \textit{Blobs} noises consist of algorithmically generated amorphous shapes with definite edges. As in Table\ref{synthetic}, our baseline OCL achieves significant improvement in the synthetic OOD datasets with previous methods. Furthermore, our approach COCL substantially outperforms the baseline OCL and achieves the SOTA performance in the synthetic OOD datasets.

\end{document}